\documentclass{article}
\usepackage[preprint]{neurips_2026}
\usepackage[utf8]{inputenc}
\usepackage[T1]{fontenc}
\usepackage{hyperref}
\usepackage{url}
\usepackage{booktabs}
\usepackage{amsfonts}
\usepackage{amsmath}
\usepackage{nicefrac}
\usepackage{microtype}
\usepackage{tabularx}
\usepackage{hyperref}
\usepackage{booktabs}
\usepackage{array}
\usepackage{xcolor}
\renewcommand{\arraystretch}{1.35}
\usepackage{microtype}
\microtypesetup{protrusion=true, expansion=true}
\usepackage[table]{xcolor}
\usepackage{graphicx}
\usepackage{algorithm}
\usepackage{algorithmic}
\usepackage{multirow}
\usepackage{makecell}
\usepackage{adjustbox}
\usepackage{subcaption}
\usepackage[most]{tcolorbox}
\usepackage{wrapfig}
\usepackage[most]{tcolorbox}   
\usepackage{natbib}
\usepackage{wrapfig}
\usepackage{natbib}
\usepackage{mathtools}
\usepackage[capitalise,noabbrev]{cleveref}

\usepackage{tikz}
\usetikzlibrary{calc}
\definecolor{selcol}{RGB}{29,158,117}
\definecolor{graycol}{RGB}{180,178,169}
\definecolor{amrcol}{RGB}{239,159,39}
\definecolor{redcol}{RGB}{226,75,74}
\definecolor{anscol}{RGB}{13,110,71}
\definecolor{wrncol}{RGB}{163,45,45}
\definecolor{dashrule}{gray}{0.55}

\newcommand{\methodname}[1]{{Robust-TO}}
\newcommand{\myhdash}{%
  \noalign{\vspace{1pt}}%
  \arrayrulecolor{dashrule!60}\hline\arrayrulecolor{black}%
  \noalign{\vspace{1pt}}%
}
\newcommand{\lblcell}[2]{%
  \begin{tabular}[t]{@{}l@{}}\textbf{#1}\\[1pt]\textbf{#2}\end{tabular}%
} 
\Crefname{section}{Section}{Sections}
\Crefname{table}{Table}{Tables}
\crefname{section}{Sec.}{Secs.}
\crefname{table}{Tab.}{Tabs.}
\crefname{figure}{Fig.}{Figs.}
\crefname{appendix}{Sec.}{Secs.}
\crefname{equation}{Eq.}{Eqs.}

\title{Confidence-Aware Tool Orchestration for \\Robust Video Understanding}

\author{%
    Yangfan He\textsuperscript{1~2} \quad\quad
    Yujin Choi\textsuperscript{1~3} 
    \quad\quad Jaehong Yoon\textsuperscript{1}\thanks{Corresponding Author} \\[1ex]
  \textsuperscript{1}NTU Singapore  \quad
  \textsuperscript{2}University of Minnesota, Twin Cities  \quad
  \textsuperscript{3}UNIST \\[1ex]
  \\
  Project Page: \textcolor{magenta}{\url{https://rova-v2.github.io/}}
}

\begin{document}
\maketitle

\begin{abstract}
Video reasoning language models implicitly assume that every input frame is equally reliable. This leads to what we term the \textit{Blind Trust Problem}: under realistic perturbations such as motion blur, glare, or occlusion, frontier video reasoning models can suffer 15-30\%p accuracy drops on real-world embodied benchmarks, while remaining unaware that their visual evidence has been degraded. To address this challenge, we propose \methodname{}, an agentic video understanding framework that explicitly integrates per-frame trustworthiness into every stage of reasoning. \methodname{} organizes heterogeneous visual perception tools under a unified evidence interface. Each tool receives a sub-query derived from the original question and a set of trustworthy frames selected by the reliability-relevance score. It returns evidence in a shared format: a concrete prediction (e.g., a bounding box, motion trajectory, recognized text, or action label), temporal grounding, and a calibrated reliability score. During reasoning, these calibrated scores guide evidence weighting in a three-tier synthesis process (high/medium/low) and define a confidence-cost GRPO reward that jointly optimizes correctness, evidence reliability, and efficiency. On two video reasoning benchmarks spanning eight tasks, \methodname{} achieves 56.4\% average accuracy on clean inputs, surpassing the strongest open-source baseline by 10.6\%p and outperforming Gemini-2.5-Pro (46.2\%). Under five realistic corruption types, \methodname{} maintains 54.3\% average accuracy, 5.8\%p above the strongest open-source baseline, while exhibiting the smallest clean-to-corrupted accuracy drop among all compared methods.
% Our project page is available at: \href{https://rova-v2.github.io/}{https://rova-v2.github.io/}
\end{abstract}

\section{Introduction}
\label{sec:intro}

When asked which car ran the red light in a video with a smudged windshield and motion blur, a careful observer does not respond immediately. 
They first decide which moments are worth trusting, lean on the clearest evidence, and re-examine ambiguous segments only if they remain uncertain.  
The decision proceeds in stages: first, a quick scan identifies interpretable segments; next, a focused examination inspects relevant details (e.g., faces, license plates, traffic-light state); and only if these are insufficient, the model deliberately revisits challenging frames with a clearer sense of what to seek. 
Contemporary video large language models (Video-LLMs)~\citep{li2024llavaonevision, wang2024qwen2vl, bai2025qwen25vl, lin2023videollava} largely omit these processes. They typically rely on uniform frame sampling~\citep{bai2025qwen25vl, li2024llavaonevision, wang2024qwen2vl}, encode the sampled frames through a vision backbone, and produce answers without explicitly evaluating whether the underlying visual evidence is sufficiently reliable to justify the prediction. This design collapses three fundamentally distinct capabilities into a one-step feed-forward prediction.

We formalize this implicit assumption as the \textit{Blind Trust Problem}: every frame is treated as equally informative, every perception output as equally reliable, and the model's confidence in its answer is decoupled from the visual conditions that produced it. Its cost is well documented and silent: recent benchmarks~\citep{li2024rbench, zhang2024corruptions, agarwal2025mvtamperbench,he2026video} show frontier video reasoning models losing 15-30\%p on UrbanVideo under common corruptions, while their self-reported confidence remains largely unchanged. Zhang et al.~\citep{zhang2024multitrust} highlight that scaling parameters and data alone do not necessarily lead to robustness.
In safety-critical settings such as forensic video analysis, surveillance review, or post-hoc autonomous driving analysis, this silent failure is precisely the mode that must be eliminated.
% \uzn{In the contribution, you wrote that [we identified blind trust problem], but it only defined in this paragraph, and based on your writing, it seems like previous works}

% \jy{This paragraph is almost the same as the Abstract; the intro should provide a more comprehensive presentation of our core idea and method. Rewrite the explanation more sharply and comprehensively based on the formal process (stages) mentioned in Fig2  and method section}
To address this problem, we introduce Robust-TO, an agentic video understanding framework that performs robust reasoning on real-world videos through adaptive visual tool use guided by frame-wise reliability estimates. The pipeline proceeds in three stages (see~\cref{fig:overview}): \textbf{(1)~Frame Selection via Quality Profiling}: we design a parameter-free \texttt{assess\_quality} tool (see~\cref{tab:tools}) that characterizes each frame’s degradation in terms of blur, brightness deviation, and occlusion, producing a disturbance profile. This profile captures both the \textit{dominant} corruption type and its \textit{severity}. Frames are then jointly scored by reliability and query relevance, filtering out corrupted yet query-relevant distractors, and retaining the top-$K$ trustworthy frames for downstream perception. \textbf{(2)~Confidence-Guided Tool Routing}: To obtain fine-grained perceptual evidence that addresses each distinct aspect of the query, we decompose the input query into atomic sub-queries and route each sub-query to the perception tool best suited to the dominant corruption observed in the selected frames. Every tool call returns a \texttt{(result, confidence)} pair, where confidence is computed as the product of the tool's intrinsic certainty and the estimated reliability of its input frames, thereby down-weighting tool outputs derived from degraded inputs during reasoning. \textbf{(3)~Video Reasoning with Reliability-Aware Evidence}: \texttt{(result, confidence)} pairs collected across all sub-query tool calls are grouped into three reliability tiers (high/medium/low) to synthesize the final answer; high-tier evidence drives the conclusion, medium-tier one is retained only if consistent, and low-tier one is considered only when no stronger evidence is available, with residual uncertainty explicitly reported in the final answer.

The host VLM is trained end-to-end with Group Relative Policy Optimization (GRPO)~\citep{shao2024deepseekmath}. The training reward combines four signals: (i)~\textit{correctness reward} that scores answer accuracy; (ii)~\textit{confidence-cost reward} that encourages high-confidence outputs while penalizing expensive tool calls on degraded frames; (iii)~\textit{sub-query efficiency reward} that penalizes both over- and under-decomposition of the query, anchored to a VLM-estimated target count; and (iv)~\textit{format reward}.
% Specifically, the confidence-cost reward grounds the training signal in per-frame visual quality through the unified \texttt{(result, confidence)} interface, encouraging the agent to match tools to the available visual evidence rather than defaulting to the most expensive perception primitive.

We demonstrate the effectiveness of the proposed Robust-TO on UrbanVideo-Bench~\citep{zhao2025urbanvideo} and VSI-Bench~\citep{yang2025thinking}, spanning eight tasks under both clean and corrupted conditions generated via RoVA~\citep{he2026video}.
On clean benchmarks, Robust-TO with Qwen3-VL-7B achieves 56.4\% average accuracy, surpassing Gemini-2.5-Pro (46.2\%) and the supervised fine-tuned Qwen2.5-VL-7B (45.8\%).
On corrupted UrbanVideo-Bench, Robust-TO achieves an average accuracy of 54.3\%, outperforming the strongest open-source baseline, Video-R1~\citep{feng2025videor1}, by 5.8\%p and the best proprietary model, Gemini-2.5-Pro, by 16.2\%p.
Moreover, the adaptive key-frame selector reduces the average number of processed frames by 35\% (from 32 to 20.7) and cuts per-sample inference time by over 35\%, while simultaneously improving accuracy by 1.6\%p (\cref{tab:ablation_kfe}). 
% In addition, \methodname{} shows the smallest clean-to-corrupted accuracy drop, with 14.3\% and 51.6\% less relative degradation than Video-R1 and Gemini-2.5-Pro, respectively.
These results highlight that the adaptive visual tool use in~\methodname{}, guided by per-frame reliability estimates, benefits both corrupted and clean videos. 

% Overall, our contributions are as follows:~\jy{I don't have a strong opinion, but if the content of these bullets highly overlaps with abs/intro, we can consider commenting this out.} \uzn{I think these bullets can be removed}
% \begin{itemize}
% % \itemsep -0.05em
% \item We identify and quantify the Blind Trust Problem in Video-LLMs \uzn{as this is the first main contribution, we should highlight Blind Trust Problem in the main text; it only defined in the Introduction}: under realistic corruptions, frontier models lose 15-30 accuracy points while their self-reported confidence remains nearly unchanged (\cref{sec:exp:main}).
% \item We introduce Robust-TO, an agentic framework with a unified \texttt{(result, confidence)} tool interface that couples tool certainty with per-frame disturbance estimates for reliability-aware evidence acquisition and synthesis (\cref{sec:tools}).
% \item We propose a confidence-cost GRPO objective with a sub-query efficiency reward that jointly optimizes correctness, reliability, and efficiency, validated by extensive ablations (\cref{sec:reward} and \cref{sec:exp}).
% \end{itemize}
\section{Related Work}
\label{sec:related}

% \jy{go through and polish the related section carefully}
\paragraph{Video Large Language Models.}
Recent Video-LLMs extend strong image-language pretraining to the video domain by explicitly modeling temporal information. LLaVA-OneVision~\citep{li2024llavaonevision} demonstrates that a single architecture can transfer image instruction tuning to video; InternVL~\citep{chen2024internvl}, Qwen2/2.5-VL~\citep{wang2024qwen2vl,bai2025qwen25vl} introduce dynamic FPS sampling, window attention, and absolute time embeddings to support hour-long inputs; earlier work such as Video-LLaVA~\citep{lin2023videollava} unifies image and video representations using a shared projector. These models achieve top results on Video-MME~\citep{fu2024videomme}, MVBench~\citep{li2024mvbench}, EgoSchema~\citep{mangalam2023egoschema}, and NExT-QA~\citep{xiao2021nextqa}. They all assume the visual signal is clean: frames are uniformly or densely sampled, and the pipeline never questions whether a given frame deserves attention. \methodname{} inherits this design as a limiting cas when frames are clean, it works exactly the same; when they are not, noise at the frame level alters the entire reasoning path.

\paragraph{Agentic and Tool-Augmented Video Reasoning.}
Researchers have also explored video understanding as iterative information acquisition. ReAct~\citep{yao2023react} and Toolformer~\citep{schick2023toolformer} establish the paradigm of interleaving reasoning traces with tool calls; ToolLLM~\citep{qin2024toolllm} scales this idea to thousands of APIs. In the video domain, VideoAgent~\citep{wang2024videoagent} uses an LLM controller with a CLIP retriever~\citep{radford2021clip} and a captioner to iteratively select frames. A memory-augmented variant~\citep{fan2024videoagent} adds temporal and object-level memory, while Graph-VideoAgent~\citep{liu2025graphvideoagent} maintains an explicit entity-relation graph. These systems show that selectively gathering evidence outperforms dense encoding, but their tool interfaces only report what was found, not how reliably. Concretely, when an LLM controller calls a CLIP retriever, it cannot tell whether a high‑similarity frame is informative or whether both the query and the frame are overwhelmed by noise. \methodname{} closes this gap by providing explicit (output, confidence) pairs, treating input quality as a direct reasoning signal for the LLM rather than burying it inside retriever scores.
\paragraph{Reinforcement Learning for Multimodal Reasoning.}
Group Relative Policy Optimization (GRPO) was introduced in DeepSeekMath~\citep{shao2024deepseekmath} as a memory-efficient alternative to PPO~\citep{schulman2017ppo}. It eliminates the value network by estimating advantages from group-normalized rewards. DeepSeek-R1~\citep{guo2025deepseekr1} shows that rule-based GRPO can elicit reflection and verification without any SFT cold-start. In the video domain, Video-R1~\citep{feng2025videor1} adds a temporal-order auxiliary reward; DeepVideo-R1~\citep{park2025deepvideor1} reformulates GRPO as advantage regression with difficulty-aware augmentation; and VideoChat-R1~\citep{li2025videochatr1} applies GRPO to spatio-temporal grounding. These rewards, however, say nothing about how the answer was reached. \methodname{} augments the correctness signal with a confidence-cost term and a question-adaptive sub-query term. Both terms produce useful gradients even when the correctness signal is ambiguous, and both directly tie back to visual quality through the unified confidence.

\section{\methodname{}: Robust Video Understanding with Tool Orchestration}
\label{sec:method}
% \methodname{} tackles the Blind Trust Problem (§\ref{sec:intro}) by making per-frame trustworthiness a first-class reasoning signal. The key idea is to condition tool selection on input degradation: the framework identifies each frame's dominant corruption mode and routes sub-queries to the most reliable tool, returning a confidence score that blends tool certainty with frame quality. Three co-designed components realize this: a confidence-reporting tool interface (§\ref{sec:tools}) with a unified (result, confidence) contract; a disturbance-aware pipeline (§\ref{sec:pipeline}) that profiles, selects, routes, and synthesizes evidence with confidence weighting; and a confidence-cost GRPO reward (§\ref{sec:reward}) that trains the host VLM end-to-end to balance correctness, efficiency, and reliability. The pipeline is identical at training and inference time.
 
%-----------------------------------------------------------------

% \subsection{Setup and Tool Interface}
% \label{sec:tools}
%-----------------------------------------------------------------
\subsection{Blind Trust Problem}
\begin{wrapfigure}{r}{0.5\textwidth} 
    \centering
    \vspace{-15pt} 
    \includegraphics[width=0.5\textwidth]{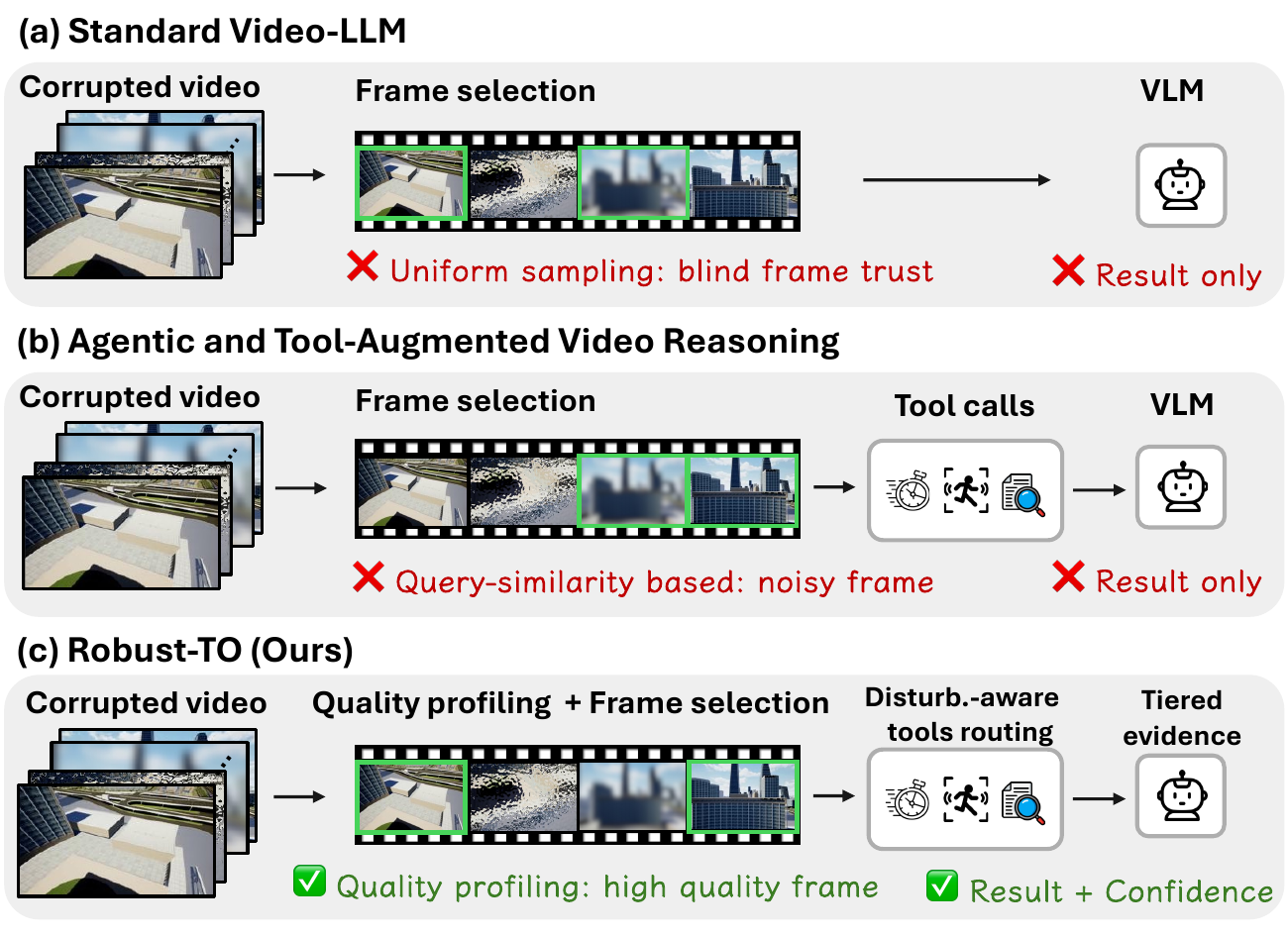}
    \vspace{-15pt}
    \caption{Comparison of video reasoning pipelines under corrupted video. }
    \label{fig:baseline}
    \vspace{-15pt}
\end{wrapfigure}
% \textcolor{red}{
Real-world videos are rarely pristine, as motion blur, glare, low-light noise, and occlusions frequently degrade visual quality. However, modern Video-LLMs typically process frames under the implicit assumption that they are equally reliable. This blind trust is harmful because degraded frames enter the reasoning process as evidence, and their corrupted visual signals can distort the final prediction.
% } 
% \uzn{15-30\% degradation is repeated three times; just stating equally trust frames can contain corrupted/noisy samples is enough}\uzn{emphasize Blind trust problem / motivation}\jy{this is important}

As shown in~\cref{fig:baseline}, existing methods either select frames uniformly or by query similarity alone, both of which can admit corrupted frames and lead to incorrect answers. 
% \textcolor{red}{this degradation is pervasive across all model families and scales: every baseline - proprietary or open-source - suffers a substantial clean-to-corrupted accuracy drop, while \methodname{} achieves both higher clean accuracy and the smallest degradation gap, confirming that the Blind Trust Problem demands a principled solution beyond mere scaling.}
\methodname{} addresses this by selecting clean frames based on both quality profiling and query similarity, even under corrupted video conditions. Moreover, it provides a confidence score alongside each answer, quantifying how much the supporting evidence can be trusted, which is particularly valuable when the input video is corrupted.

\begin{figure}
    \centering
    \includegraphics[width=\linewidth]{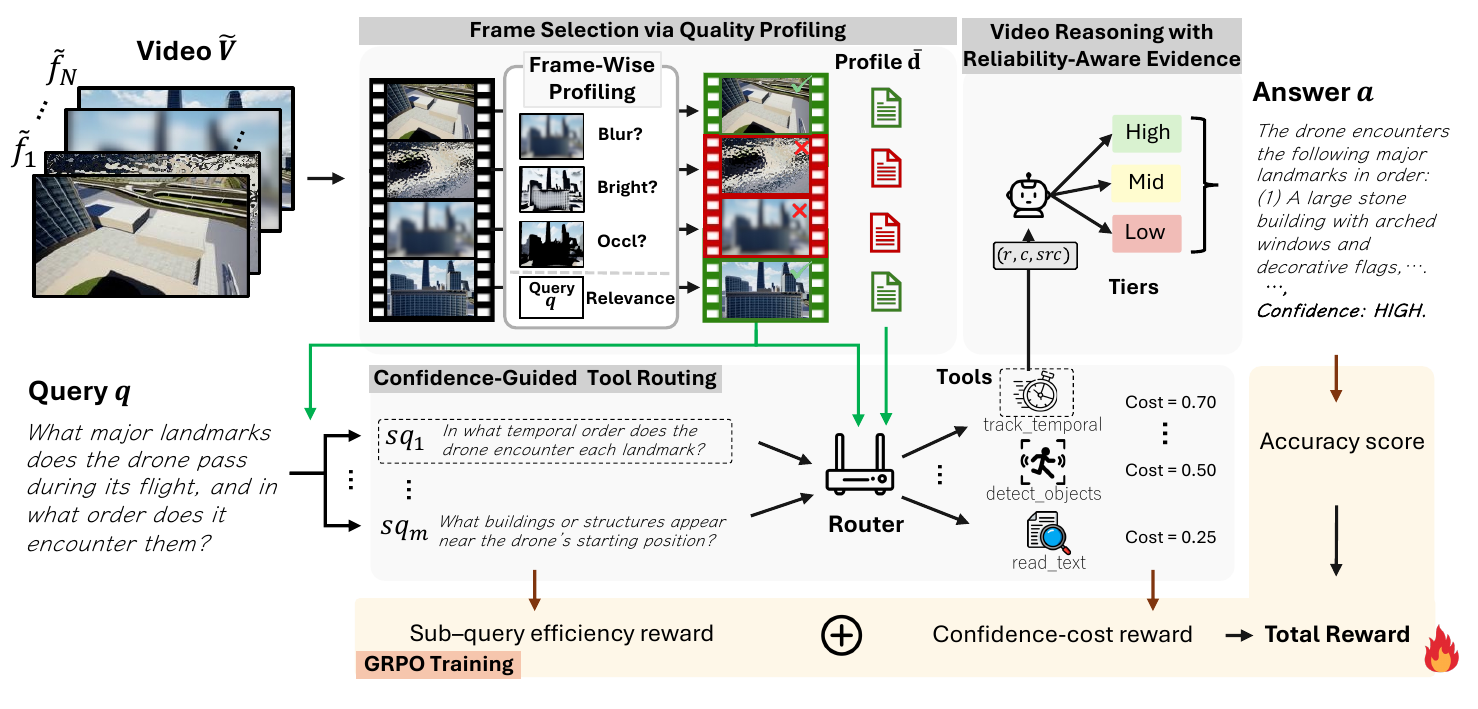}
    \vspace{-0.2in}
    \caption{Overview of \methodname{}. Given a real-world video $\tilde{\mathcal{V}}$ and a query $q$, the host VLM first profiles each frame's quality, then selects the most reliable frames. It decomposes $q$ into atomic sub-queries and routes each to the perception tool best matched to the dominant corruption. Finally, it predicts the answer by grouping evidence into reliability tiers. The host VLM is trained end-to-end with GRPO using a reward that combines accuracy, efficiency, and confidence-cost trade-off.}
    \label{fig:overview}
    \vspace{-0.1in}
\end{figure}
\subsection{Problem Setting}
Let $\mathcal{V} = \{f_1, \dots, f_N\}$ denote a clean video of $N$ frames.
% \jy{Motivated by the real-world perturbations, we assume that the learner observes video streams}
Motivated by real-world perturbations, we assume that the learner observes video streams
$\tilde{\mathcal{V}} = \{\tilde{f}_1, \dots, \tilde{f}_N\}$, where each frame may be corrupted by unknown degradations with some probability:
\begin{equation}
  \tilde{f}_i = \mathcal{D}_i(f_i,\, \delta_i), \qquad
  \mathcal{D}_i : \mathcal{X} \times [0,1] \to \mathcal{X}, \qquad
  i = 1, \dots, N,
  \label{eq:corruption}
\end{equation}
with per-frame severity $\delta_i \in [0,1]$
($\delta_i{=}0$: no degradation; $\delta_i{=}1$: complete information loss).
Neither the corruption family $\mathcal{D}$ nor the severity schedule
$\delta$ is known to the system in advance, and different frames may be affected by distinct disturbance types (e.g., motion blur, glare, or partial occlusion).

Given a text query $q$ over $\tilde{\mathcal{V}}$, the goal is to produce a correct answer $a$ while satisfying two requirements: \emph{(i)}~infer the trustworthiness of each frame from the observed pixels alone, without supervision; and \emph{(ii)}~condition evidence acquisition (i.e., the selection of input frames and the dispatch of perception tools that gather evidence from them) and reasoning on these inferred reliability signals, so that conclusions are never grounded in frames whose visual information has been severely degraded by corruption.
When $\delta_i = 0$ for all $i$, the video is clean, and the pipeline becomes identical to standard video reasoning with no additional overhead. 
To achieve \emph{adaptive} robustness under heterogeneous and unknown corruption, we propose \methodname{}, a novel framework that identifies and estimates frame-level degradations at inference time while minimizing their disruptive effects on downstream reasoning, illustrated in~\cref{fig:overview}.
 
%-----------------------------------------------------------------
\subsection{Frame Selection via Quality Profiling and Adaptive Tool Use for Video Reasoning}
\label{sec:pipeline}

%------ Frame-level quality profiling ------
\paragraph{Frame Selection via Quality Profiling.}
Since both the corruption type and severity are unknown a priori (\cref{eq:corruption}) and can vary across frames, a vision language model (VLM) cannot reliably select trustworthy frames or appropriate tools without first assessing frame degradation. We therefore estimate the quality of each frame upfront, converting latent corruption into an explicit per-frame signal for downstream frame selection and tool routing. We define quality profiling as the process of estimating a per-frame disturbance score \(d(f_i)\) (for simplicity, we use $f_i$ to denote the observed frame $\tilde{f}_i$) alongside its component-wise degradation indicators to characterize the type and severity of corruption in each frame. We focus on blur, brightness deviation, and occlusion because they cover the dominant sources of visual unreliability in real-world video capture, including motion or defocus blur, glare, and under- or overexposure, and physical occlusion. Moreover, each can be estimated efficiently from signal-level statistics without training data or reference frames. Given the host VLM as a controller, we first invoke \texttt{assess\_quality} (the parameter-free image-quality-assessment tool in our tool set; see \cref{tab:tools}) on each frame to compute
\begin{equation}
  d(f_i) \;=\; \, \operatorname{mean}\left(d_{\mathrm{blur}}(f_i),
             d_{\mathrm{bright}}(f_i),
              \,d_{\mathrm{occl}}(f_i)\right),
  \label{eq:disturbance}
\end{equation}
where $d_{\mathrm{blur}}$ quantifies spatial sharpness via the inverse Laplacian variance (blurrier frames score higher), $d_{\mathrm{bright}}$ measures illumination distortion, assigning higher scores to frames that are either too dark or too bright, and $d_{\mathrm{occl}}$ captures the fraction of the frame lacking informative edge structure, estimated from Sobel-magnitude statistics. Please see detailed formulations in \cref{appendix:disturbance}. We min-max normalize each component across $\tilde{\mathcal{V}}$ so that blur, brightness, and occlusion contribute on a comparable scale. Beyond the aggregate disturbance score $d(f_i)$, the component-wise values $(d_{\mathrm{blur}}, d_{\mathrm{bright}}, d_{\mathrm{occl}})$ indicate which degradation is most prominent in each frame. The controller can then use this disturbance profile to prefer frames and tools that are more reliable under the corresponding degradation.

Given the disturbance scores, the controller calls the \texttt{select\_frames} tool, which ranks candidate frames using a reliability-relevance score. The tool assigns a score $s(f_i)$ to each candidate frame $f_i$:
% Candidates are scored by
% \begin{equation}
% \begin{split}
% s(f_i) = &\;\mathbf{1}\bigl(1-d(f_i) \ge \theta_{\text{rel}}\bigr)
%           \;\cdot\;
%           \mathbf{1}\bigl(\mathrm{sim}(\phi(f_i),\psi(q)) \ge \theta_{\text{sim}}\bigr) \\
%           &\;\cdot\;
%           \bigl(1-d(f_i)\bigr)
%           \;\cdot\;
%           \mathrm{sim}\!\bigl(\phi(f_i),\;\psi(q)\bigr),
% \qquad \text{for } f_i \in \mathcal{F},
% \end{split}
% \label{eq:selection}
% \end{equation}
% where $\theta_{\text{rel}}, \theta_{\text{sim}}$ are hyperparameters and $\mathbf{1}(\cdot)$ is the indicator function.
\begin{equation}\label{eq:selection}
  s(f_i) =
  (1-d(f_i)) \cdot \mathrm{sim}(\phi(f_i), \psi(q)),
  \quad f_i \in \mathcal{F}.
\end{equation}
Here, $1-d(f_i)$ estimates frame reliability, while $\mathrm{sim}(\phi(f_i), \psi(q))$ measures relevance to the query via cosine similarity. $\mathcal{F}$ denotes the set of valid
frames whose reliability and query relevance both exceed their respective
thresholds. Next, it returns the top-$K$ remaining frames for downstream perception, where $K \in [4,12]$ is adaptively chosen by the host VLM according to query complexity (See sensitivity analysis in~\cref{sec:exp:abl}). By prioritizing frames that are both visually trustworthy and query-relevant, this selection strategy provides a reliable foundation for robust downstream reasoning.

% Frames that fail either threshold receive $s(f_i)=0$ and are discarded.
% The multiplicative coupling remains crucial: as $(1-d(f_i))$ approaches $0$ under severe degradation, heavily corrupted frames are suppressed even when query relevance is high, preventing unreliable frames from being accidentally promoted into the top-$K$.
% The number of selected frames $K\in[4,12]$ is then adaptively determined by the host VLM according to query complexity (see sensitivity analysis in \cref{sec:exp:abl}).
%------ Disturbance-aware tool routing ------
\paragraph{Confidence-Guided Tool Routing.}
After high-reliability frames are selected, the host VLM determines which perception tools should process them. 
Complex queries often require multiple perceptual skills. For example, answering whether an object moves after appearing in a scene requires both spatial localization and motion analysis. Since different perception tools specialize in different skills and exhibit different robustness to blur, illumination degradation, and occlusion, the controller first decomposes the query $q$ into atomic sub-queries $\{sq_1,\ldots,sq_m\}$ conditioned on the original question and the selected frames. Each sub-query targets a single perceptual primitive, such as object localization, motion analysis, OCR, or attribute recognition, and is generated through in-context instructions without a dedicated parser.

% Because a complex query $q$ often conflates multiple perceptual demands - e.g.\ localizing an object \emph{and} judging its motion - a single tool call is both brittle and inefficient. 
% Therefore, it splits $q$ into atomic sub-queries $\{sq_1, \dots, sq_m\}$, a process that depends jointly on the original $q$ and the visual content of the identified trustworthy frames, each targeting one perceptual primitive (spatial grounding, temporal reasoning, attribute recognition, etc.), via in-context instructions without any dedicated parser. 
Routing then follows a two-stage \textit{plug-and-play} protocol that is agnostic to the specific tool library. First, the semantic type of each sub-query determines the candidate tools. For example, spatial queries are matched to \texttt{detect\_objects}, while temporal queries are matched to \texttt{track\_temporal} or \texttt{recognize\_action} (see \cref{tab:tools}).
Second, the averaged disturbance profile $\bar{\mathbf{d}}=(\bar d_{\mathrm{blur}}, \bar d_{\mathrm{bright}}, \bar d_{\mathrm{occl}})$ selects the most reliable candidate by identifying the dominant corruption type: when blur is predominant, the controller favors \texttt{caption\_frame}, which is more tolerant to spatial degradation, over \texttt{detect\_objects}, which depends on sharp boundaries; when brightness distortion dominates, it prioritizes tools that are more robust to extreme illumination.

Let $\mathbf{F}_j \subseteq \mathcal{F}$ denote the set of selected high-reliability frames provided to the $j$-th tool call for sub-query $sq$, with $j$ indexing tool calls globally. Each call returns a result-confidence pair $(r_j,c_j)$ in a shared output format, where $r_j$ is the tool's perception output (e.g., a bounding box, recognized text, action label, or caption) and $c_j$ is the confidence score associated with that output.
\begin{equation}
  (r_j, c_j) = T_j(\mathbf{F}_j, sq), \qquad
  c_j \;=\; \underbrace{c_j^{\mathrm{intrinsic}}}_{\text{tool self-assessment}} \;\times\; \underbrace{\rho(\mathbf{F}_j)}_{\text{input reliability}},
  \label{eq:confidence}
\end{equation}
where $c_j^{\mathrm{intrinsic}}\in[0,1]$ is a self-assessment returned by the tool itself (e.g., the mean detection score or token log-probability; see \cref{tab:tools}), and $\rho(\mathbf{F}_j)$ is the input reliability, defined as a conservative \emph{mean} of the per-frame reliabilities $1-d(f)$:
\begin{equation}
  \rho(\mathbf{F}_j) \;=\; \frac{1}{\lceil n/3\rceil}\!\!\sum_{f \in \mathbf{F}_{j,\mathrm{lowest}}}\!\!\bigl(1 - d(f)\bigr), \qquad n = |\mathbf{F}_j|,
  \label{eq:rho}
\end{equation}
where $\mathbf{F}_{j,\mathrm{lowest}}$ is the subset of $\lceil n/3\rceil$ frames in $\mathbf{F}_j$ with the lowest reliability. This penalizes tool inputs that contain severely corrupted frames and prevents a small number of clean frames from masking unreliable visual evidence. Multiplying the tool confidence $c_j$ by the frame reliability $\rho(\mathbf{F}_j)$ allows the model to avoid overconfident reasoning based on unreliable visual evidence. We define each evidence item as a tuple $(r_j, c_j, \mathbf{F}_j, \{d(f)\}_{f \in \mathbf{F}_j})$, where $r_j$ is the perception-tool output, $c_j$ is its confidence score defined in \cref{eq:confidence}, $\mathbf{F}_j$ is the set of source frames used by the $j$-th tool call, and $\{d(f)\}_{f \in \mathbf{F}_j}$ are the corresponding disturbance scores. Thus, each evidence item jointly records the tool output, its calibrated confidence, the source frames used to produce it, and the corresponding disturbance scores.
Thus, each evidence item fully tracks the tool output, its confidence, the identity of its source frames, and their disturbance scores. The complete evidence set accumulated over all sub-query tool calls is then given by $\mathcal{I} = \{(r_j, c_j, \mathbf{F}_j, \{d(f)\}_{f \in \mathbf{F}_j})\}_j$, which serves as the basis for the reliability-aware video reasoning stage described later.

\paragraph{Video Reasoning with Reliability-Aware Evidence.} 
After all sub-queries are processed, the host VLM produces the final answer $a$ by integrating $\mathcal{I}$, the complete evidence set gathered across all sub-query tool calls. Whereas each tool call yields only a local result $r_j$, the answer $a$ is produced only at this stage, after reliability-aware integration over the full evidence set. Here, we note that duplicate tool calls may occur across sub-queries. We disambiguate them with a unique global index $j$ for each tool call, while each evidence item records its source sub-query $sq_k$ and invoked tool $T_j$. 
% Each evidence item is evaluated using its confidence $c_j$ (\cref{eq:confidence}) and the disturbance scores of its source frames, $\{d(f)\}_{f \in \mathbf{F}_j}$.

Based on these signals, the host VLM groups evidence into three reliability tiers: \textit{high}, \textit{medium}, and \textit{low}. It first infers a preliminary conclusion from high-reliability evidence (\textit{high}), then evaluates each medium-reliability item (\textit{medium}) against this conclusion through in-context reasoning. Medium-reliability evidence is retained only when it agrees with the conclusion, while evidence that points to a different answer or reports an inconsistent attribute is discarded as unreliable. Low-reliability evidence \textit{(low)} is used only as a fallback when no high-reliability evidence is available, and the answer explicitly marks the remaining uncertainty. Thus, reliable evidence determines the prediction, and uncertain evidence can support but cannot change the conclusion. On clean videos, the reliability criterion naturally admits nearly all evidence, so the method reasons over the full evidence set.%-----------------------------------------------------------------
\subsection{Confidence-Cost Trade-off Reward for GRPO Training}
\label{sec:reward}
We train the host VLM with GRPO to perform confidence-guided routing and evidence integration, using a reward that balances confidence and efficiency. The core intuition is that reliable evidence (high confidence) is valuable, but obtaining it may require expensive tools (high cost). The reward encourages the model to seek high-confidence outputs while penalizing unnecessary tool expenditures, thus naturally balancing the confidence-cost trade‑off. We compute all reward components on a shared single-rollout trajectory $\tau$, defined over the full episode from frame selection, sub-query decomposition, and tool routing to final evidence integration.

\paragraph{Confidence-Cost Reward.}
For each tool $T_j$ called on sub-query $sq$ with output $(r_j, c_j)$, we define:
\begin{equation}
  R_{\mathrm{cc}}(c_j, T_j) \;=\; c_j \;-\; \lambda \cdot \mathrm{cost}(T_j),
  \label{eq:rcc}
\end{equation}
where $\mathrm{cost}(T_j)$ denotes the tool cost, defined as the wall-clock runtime normalized by the runtime of \texttt{caption\_frame} (see \cref{tab:tools}). 
% \textcolor{brown}{We set $\lambda = 0.5$ so that the penalty for the most expensive tool (max cost $0.7$) equals a medium‑confidence score of $0.35$; failed calls (i.e., tool invocations that return an error or produce no valid output) get $c_j = 0$}~\jy{this is an implementation detail, can we move this explanation to the appendix?}. 
For a trajectory $\tau$ with $N_{\mathrm{call}}$ tool calls, the total reward averages the per‑call rewards:
\begin{equation}
  R_{\mathrm{cc}}^{\mathrm{total}}(\tau) \;=\;
    \frac{1}{N_{\mathrm{call}}} \sum_{k=1}^{N_{\mathrm{call}}}
    R_{\mathrm{cc}}\!\bigl(c_{j_k},\, T_{j_k}\bigr).
  \label{eq:rcctotal}
\end{equation}
Here, $j_k$ denotes the index of $k$-th tool call, and $c_{j_k}$ is the confidence returned by tool $T_{j_k}$. %; $\tau$ is the full episode trajectory, as defined above.}

\paragraph{Sub-Query Efficiency Reward.}
Let  $m^*$ be the question-dependent optimal number of sub-queries. We want the VLM to decompose $q$ into the right number of sub-queries $m$, avoiding both information gaps ($m < m^*$) and wasteful calls ($m > m^*$). We estimate  $m^*$ using a separate, frozen off-the-shelf VLM $\pi_{\mathrm{est}}$ (not the policy VLM itself) once per question; $\pi_{\mathrm{est}}$ is a pretrained VLM that we prompt to predict the minimal number of sub-queries needed to answer $q$ reliably. Here, $R_{\mathrm{min\text{-}sq}}$ discourages excessive sub-query decomposition, while $R_{\mathrm{qual}}(\tau)$ rewards trajectories that use a sufficient number of tools by computing the average tool confidence over the same trajectory $\tau$. The sub-query efficiency reward $R_{\mathrm{subq}}$ is then defined as the sum of these two terms, $R_{\mathrm{subq}} = R_{\mathrm{min\text{-}sq}} + R_{\mathrm{qual}}(\tau)$.
\paragraph{Total Reward.} With rewards defined for tool‑use efficiency ($R_{\mathrm{cc}}$) and for the overall decomposition strategy ($R_{\mathrm{subq}}$), we now combine them with correctness and format signals into a single training objective. The format reward $R_{\mathrm{fmt}} \in \{0,1\}$ indicates whether the model's output follows the required structure (e.g., JSON format for tool calls or final answers), {and $R_{\mathrm{acc}} \in \{-1, +1\}$ indicates whether the final answer is correct}. The four terms above are combined into a single scalar reward for each trajectory:
\begin{equation}
R_{\mathrm{total}} = R_{\mathrm{acc}} + w \left( R_{\mathrm{subq}} + R_{\mathrm{cc}}^{\mathrm{total}} + R_{\mathrm{fmt}} \right),
\label{eq:rtotal}
\end{equation}
where $w = 1/3$, and other auxiliary terms are bounded. %\uzn{is this necessary sentence? i think we don't need to state the reason why $w=1/3$} %$R_{\mathrm{acc}} \in \{-1, +1\}$, $R_{\mathrm{fmt}} \in \{0, 1\}$,

%=====================================================================
% EXPERIMENTS
%=====================================================================
\section{Experiments}
\label{sec:exp}
% We address four key questions: (1) Does \methodname{} preserve its advantage on degraded frames? (\cref{sec:exp:main}) (2) Does the confidence‑reporting interface prevent reliance on unreliable evidence? (\cref{sec:exp:abl}) (3) Does disturbance‑aware tool routing outperform simpler baselines? (\cref{sec:exp:abl}) (4) Are the confidence‑cost and sub‑query efficiency rewards essential? (\cref{sec:exp:abl})~\jy{this reference is unclear as most of them just cite sec 4.3.} By systematically ablating each component and comparing against strong baselines, we isolate the sources of \methodname{}’s corruption robustness.

\begin{table*}[t!]
    \centering
    \caption{Performance on 8 indoor and outdoor embodied spatial reasoning tasks. The baselines include popular proprietary, open-source multimodal reasoning models, video LLMs, and models fine-tuned on the same training data. Tasks include: LP (Landmark Position), CF (Counterfactual), PE (Progress Evaluation), AG (Action Generation), RDist (Relative Distance), RDir (Relative Direction), RP (Route Planning), AO (Appearance Order).}
    \label{tab:acc}
    \setlength{\tabcolsep}{4pt}
    \centering
    \footnotesize
    \renewcommand{\arraystretch}{1}
    \begin{tabular*}{\textwidth}{@{\extracolsep{\fill}}l|c|c|cccccccc@{}}
    \toprule
    & & & \multicolumn{4}{c}{\cellcolor{cyan!20}UrbanVideo-Bench} & \multicolumn{4}{c}{\cellcolor{green!20}VSI-Bench} \\
    Method & Frames & Avg. &
        \textit{LP} &
        \textit{CF} &
        \textit{PE} &
        \textit{AG} &
        \textit{RDist} &
        \textit{RDir} &
        \textit{RP} &
        \textit{AO} \\
    \midrule
    \multicolumn{11}{l}{\cellcolor[HTML]{F5F5F5}\textit{\textbf{Proprietary Models (API)}}} \\
    Qwen-VL-Max& 32        & 35.2 & 44.8 & 49.2 & 38.8 & 29.6 & 28.0 & 33.3 & 29.6 & 28.3 \\
    GPT-4o &32             & 36.0 & 36.8 & 44.7 & 34.2 & 33.8 & 37.0 & 41.3 & 31.5 & 28.5 \\
    Gemini-1.5-Flash &1fps  & 38.2 & 37.8 & 42.4 & 43.3 & 34.4 & 37.7 & 41.0 & 31.5 & 37.8 \\
    Gemini-1.5-Pro &1fps    & 40.3 & 37.4 & 46.2 & 38.8 & 31.9 & 51.3 & 46.3 & 36.0 & 34.6 \\
    \midrule
    \multicolumn{11}{l}{\cellcolor[HTML]{F5F5F5}\textit{\textbf{SOTA Reasoning Models (API)}}} \\
    OpenAI-o1 &32          & 40.4 & 34.6 & 53.3 & 39.1 & 28.0 & 39.7 & 35.8 & \textbf{52.9} & 39.8 \\
    Gemini-2.5-Pro &1fps    & 46.2 & 40.0 & \textbf{75.0} & 38.7 & 23.5 & 42.0 & 34.5 & 52.4 & 63.6 \\
    \midrule
    \multicolumn{11}{l}{\cellcolor[HTML]{F5F5F5}\textit{\textbf{Open-source Models}}} \\
    LLaVA-NeXT-Video-7B-hf &32   & 29.0 & 49.5 & 20.5 & 36.6 & 19.2 & 25.2 & 26.3 & 29.9 & 24.5 \\
    Phi-3.5-vision-instruct &32 & 30.9 & 49.2 & 34.8 & 33.2 & 15.6 & 25.4 & 26.5 & 36.9 & 25.2 \\
    Kangaroo &32                & 30.4 & 35.5 & 42.4 & 32.5 & 32.4 & 25.2 & 26.8 & 23.5 & 24.9 \\
    InternVL2-2B &32             & 27.7 & 19.3 & 45.5 & 29.2 & 20.9 & 25.1 & 25.0 & 32.6 & 23.9 \\
    InternVL2-8B &32            & 28.1 & 23.1 & 45.5 & 31.5 & 21.4 & 24.7 & 25.7 & 28.3 & 24.8 \\
    InternVL2-40B &32           & 28.0 & 23.2 & 41.7 & 32.4 & 22.3 & 24.9 & 25.7 & 29.4 & 24.5 \\
    Qwen2.5-VL-3B-Instruct &32 & 35.4 & 32.1 & 47.8 & 34.0 & 31.0 & 27.9 & 32.6 & 39.0 & 38.9 \\
    Qwen2.5-VL-7B-Instruct &32 & 33.9 & 33.3 & 21.7 & 25.0 & 27.8 & 35.8 & 39.7 & 48.8 & 38.8 \\
    Qwen2.5-VL-72B-Instruct &32 & 35.0 & 34.7 & 34.8 & 26.4 & 37.7 & 40.8 & 29.0 & 32.5 & 43.9 \\
    \midrule
    \multicolumn{11}{l}{\cellcolor[HTML]{F5F5F5}\textit{\textbf{Supervised Fine-Tuning}}} \\
    Qwen2.5-VL-3B-Instruct &32 & 40.6 & 47.7 & 33.4 & 34.8 & 39.2 & 42.6 & 42.3 & 41.2 & 43.9 \\
    Qwen2.5-VL-7B-Instruct &32 & 45.8 & 40.2 & 53.4 & 38.0 & 40.8 & 47.8 & 46.3 & 44.1 & 56.1 \\
    \midrule
    \multicolumn{11}{l}{\cellcolor[HTML]{FFE4B5}\textit{\textbf{\methodname{} (Ours)}}}  \\
    \textbf{\methodname{}} (Qwen2.5-VL-7B-Instruct) & 20.7 & 50.7 & 55.1 & 59.9 & 39.7 & 47.6 & 50.0 & 44.3 & 36.8 & 72.0 \\
    \rowcolor[HTML]{E8F5E9}
    \textbf{\methodname{}} (Qwen3-VL-7B-Instruct)   & 20.7 & \textbf{56.4} & \textbf{61.1} & {64.4} & \textbf{44.7} & \textbf{59.0} & \textbf{55.5} & \textbf{48.8} & 39.8 & \textbf{77.5} \\
    \bottomrule
    \end{tabular*}
\vspace{-0.1in}
\end{table*}

\begin{table*}[t!]

    \centering

    \caption{Accuracy on UV-Bench under each of the five PVRBench~\cite{he2026video} corruption masks (MB: Motion Blur, GN: Gaussian Noise, GL: Glare, Occ: Occlusion, LL: Low-Light) for selected models. Numbers are averaged across the four UV-Bench tasks (LP, CF, PE, AG).}

    \label{tab:per_mask_breakdown}

    \setlength{\tabcolsep}{6pt}

    \footnotesize

    \renewcommand{\arraystretch}{1}

    % \textcolor{red}{Added a dedicated ``Frames'' column so every method has consistent frame/sampling info; removed [32]/[1fps] from method names.}

    \begin{tabular*}{\textwidth}{@{\extracolsep{\fill}}l|c|c|ccccc|c}

    \toprule

    Method & {Frames} & Clean & MB & GN & GL & Occ & LL & Avg \\

    \midrule

    GPT-4o                                  & {32}   & 37.4 & 32.2 & 31.7 & 32.5 & 30.8 & 33.6 & 32.2 \\

    Gemini-2.5-Pro                          & {1fps} & 44.3 & 37.7 & 37.7 & 38.8 & 36.4 & 39.8 & 38.1 \\

    Qwen2.5-VL-7B-Instruct                 & {32} & 26.9 & 17.0 & 16.6 & 18.9 & 14.6 & 20.3 & 17.5 \\

    Qwen2.5-VL-72B-Instruct                & {32} & 33.4 & 26.3 & 26.0 & 27.2 & 24.3 & 28.7 & 26.5 \\

    \midrule

    Qwen2.5-VL-7B-Instruct (SFT)           & {32}   & 39.0 & 30.9 & 30.7 & 32.1 & 29.2 & 33.3 & 31.2 \\

    \midrule

    Video-R1 (Qwen2.5-VL-7B-Instruct)      & {32} & 43.0 & 38.5 & 38.0 & 39.5 & 37.0 & 40.5 & 38.7 \\

    Video-R1 (Qwen3-VL-7B-Instruct)        & {32} & 52.0 & 48.5 & 48.0 & 49.0 & 47.5 & 49.5 & 48.5 \\

    \midrule

    \textbf{\methodname{} (Ours)} (Qwen2.5-VL-7B-Instruct) & {20.7} & 50.6 & 47.0 & 46.5 & 47.7 & 46.1 & 48.3 & 47.1 \\

    \rowcolor[HTML]{E8F5E9}

    \textbf{\methodname{} (Ours)} (Qwen3-VL-7B-Instruct)   & {20.7} & \textbf{57.3} & \textbf{54.1} & \textbf{54.0} & \textbf{54.9} & \textbf{53.5} & \textbf{55.1} & \textbf{54.3} \\

    \bottomrule

    \end{tabular*}
\vspace{-0.2in}
\end{table*}
\subsection{Experiment Setup}
We evaluate \methodname{} under realistic video corruptions, including motion blur, glare, and occlusion. We use RoVA~\cite{he2026video} to generate degraded variants of UrbanVideo-Bench (UV-Bench)~\cite{zhao2025urbanvideo} and VSI-Bench~\cite{yang2025thinking}, enabling controlled assessment of robustness across both clean and corrupted inputs.
We instantiate \methodname{} as a training framework on two base models: Qwen2.5-VL-7B \cite{bai2025qwen25vl}, and Qwen3-VL-7B \cite{bai2025qwen3}.  Training is conducted on the video subset of the Video-R1 dataset, which covers both indoor and outdoor scenes,  using GRPO with $4\times$A100 GPUs (rollout group size of 16 for $\sim$5k steps). To prevent the policy model from exploiting the sub-query count during training, we estimate the optimal number of sub-queries using a frozen Qwen2.5-VL-7B-Instruct model.
% For evaluation, we follow the original \methodname{} benchmarks -\jy{dash} UV-Bench and VSI-Bench - and further augment them 
% with corrupted variants generated by the RoVA~\cite{he2026video} video masker, enabling systematic assessment of robustness under controlled degradations.

\begin{figure*}[t!] \centering \begin{minipage}[t]{0.53\linewidth} \vspace{0pt} \centering \sloppy \captionof{table}{Paradigm ablation on UV-Bench (Qwen3-VL-7B). Each row adds one component. P$\to$C+SQ: Perception-to-Contemplate with Sub-Query Decomposition; Contemplate is the reasoning step (\cref{appendix:prompt_decompose,appendix:prompt_synthesis}).}
\label{tab:abl_paradigm} \small 
\renewcommand{\arraystretch}{1}
\setlength{\tabcolsep}{3pt} \resizebox{\linewidth}{!}{%
\begin{tabular}{lccccc} \toprule \textbf{Reasoning Paradigm} & \textbf{Avg.} & \textbf{LP} & \textbf{CF} & \textbf{PE} & \textbf{AG} \\ \midrule Direct (R1) & 39.5 & 42.1 & 44.8 & 33.6 & 37.5 \\ P$\to$C+SQ & 42.8 & 45.7 & 48.4 & 35.9 & 41.2 \\ +Tool & 49.4 & 52.8 & 55.9 & 39.8 & 49.1 \\ +Conf & \underline{52.6} & \underline{56.0} & \underline{59.4} & \underline{41.9} & \underline{53.1} \\ \rowcolor[HTML]{E8F5E9} +GRPO & \textbf{57.3} & \textbf{61.1} & \textbf{64.4} & \textbf{44.7} & \textbf{59.0} \\ \bottomrule \end{tabular}} \end{minipage} \hfill 
\begin{minipage}[t]{0.44\linewidth} \vspace{-10pt} \centering \includegraphics[width=\linewidth]{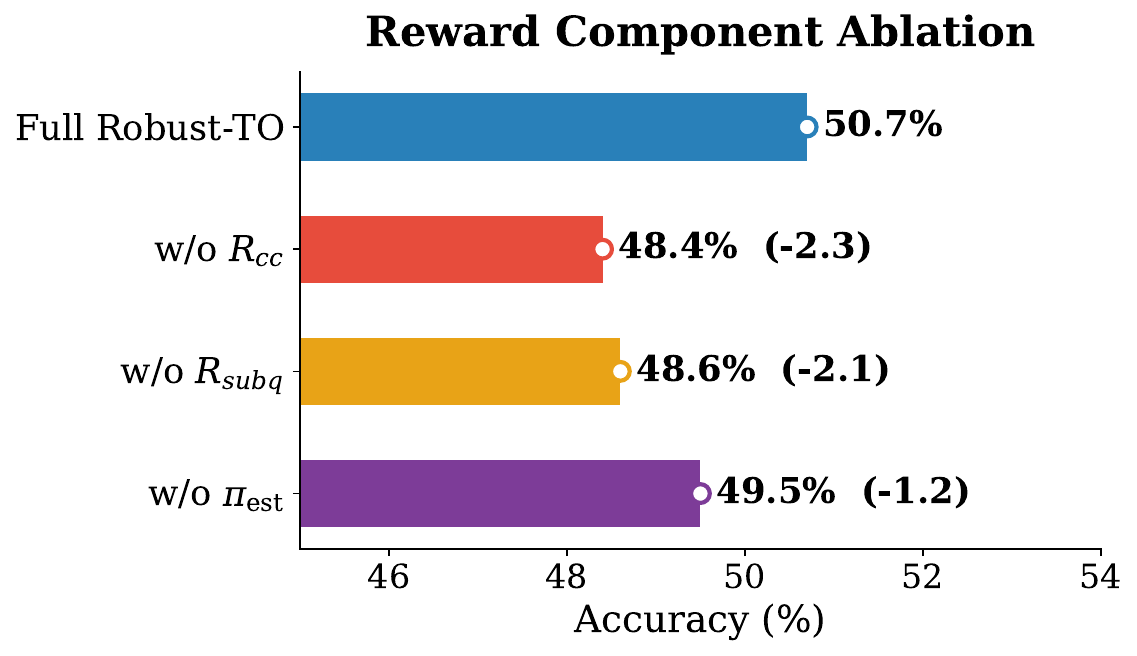} \vspace{-20pt}\captionof{figure}{Ablation of GRPO reward components on UV-Bench (Qwen2.5-VL-7B). The frozen estimator $\pi_{\mathrm{est}}$ predicts the target sub-query count $m^{*}$ using a VLM.}
\label{fig:abl_reward} \end{minipage} 
\end{figure*}

\subsection{Main Results}
\label{sec:exp:main}

\textbf{Results on Clean Dataset.}
We first evaluate \methodname{} on two challenging video reasoning benchmarks, UV-Bench \cite{zhao2025urbanvideo} and VSI-Bench \cite{yang2025thinking}, covering outdoor and indoor scenes. As shown in \cref{tab:acc}, with Qwen3-VL-7B and Qwen2.5-VL-7B backbones, \methodname{} achieves 56.4\% and 50.7\% average accuracy, respectively, outperforming GPT-4o and the corresponding base models. It achieves the best performance on 6 out of 8 tasks, with the largest gains on temporally extended tasks such as Appearance Order and Landmark Position, where reasoning requires integrating evidence from separated frames. These results show that \methodname{} improves clean-video reasoning by selecting reliable, query-relevant frames and integrating tool-assisted evidence with calibrated confidence.

\begin{wrapfigure}{r}{0.6\textwidth}
    \centering
    \vspace{-0.2cm}
    \includegraphics[width=\linewidth]{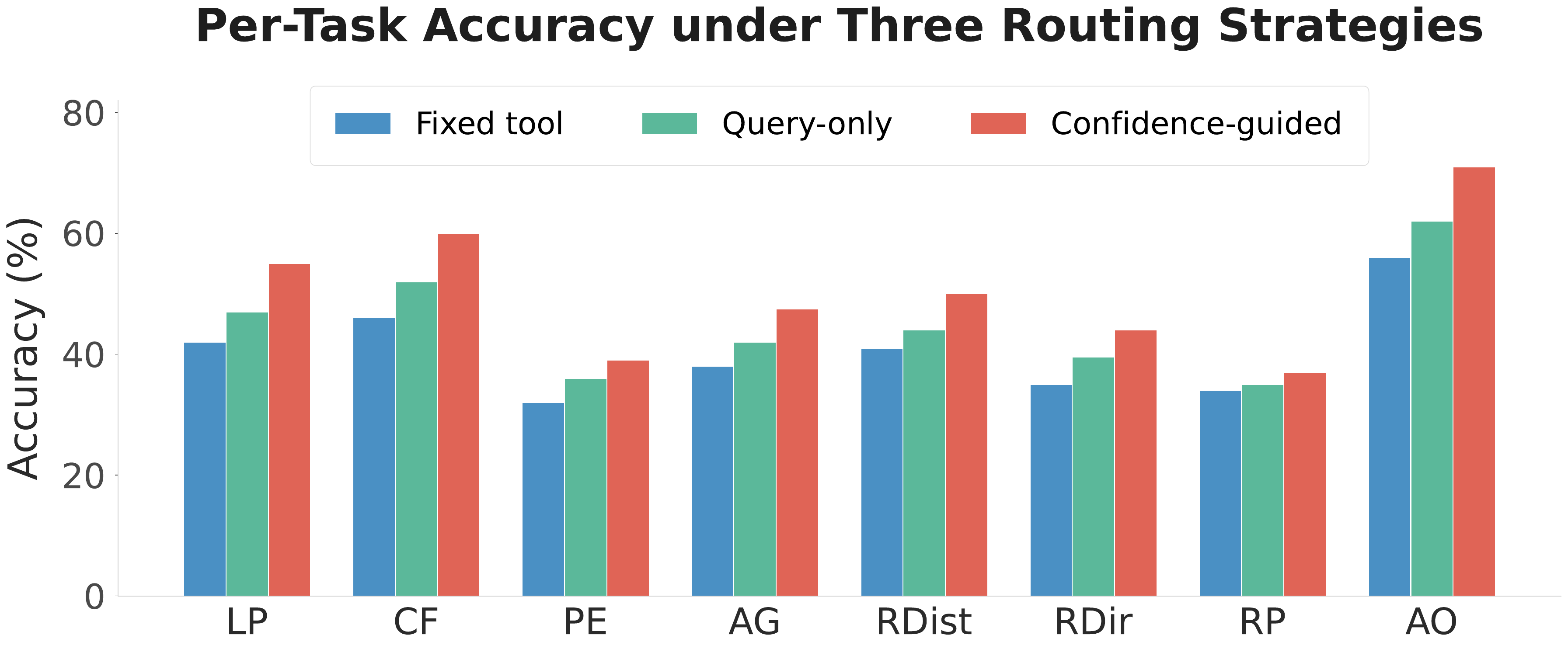}
    \caption{Ablation of Tool Routing. Per-task accuracy under three routing strategies: fixed tool, query-only, and the full confidence-guided policy.}
    \label{fig:abl_route}
    \vspace{-0.2cm}
\end{wrapfigure}
\textbf{Results on Disturbed Dataset.}
We next evaluate robustness under corrupted videos using RoVA-degraded UV-Bench, which includes five corruption types across four tasks. \methodname{} with Qwen3-VL-7B achieves 54.3\% average accuracy, substantially outperforming the best open-source baseline Video-R1 (48.5\%), the best proprietary model Gemini-2.5-Pro (38.1\%), and GPT-4o (32.2\%). \methodname{} leads on every corruption type, with particularly large gains under Occlusion and Glare. Using the same Qwen2.5-VL-7B backbone, \methodname{} also significantly improves over Video-R1 and even surpasses the much larger 72B model. Moreover, \methodname{} exhibits the smallest drop from clean to corrupted data among all methods, indicating graceful degradation under visual disturbance. With GRPO, the full \methodname{} pipeline reaches 57.3\% on the disturbed UV-Bench, while the key-frame extractor reduces the number of processed frames and inference time by 35\% with a +1.6 point accuracy gain. These results validate the central design of \methodname{}: coupling confidence with frame quality and filtering unreliable visual evidence prevents corrupted frames from dominating downstream reasoning. 
\subsection{Ablation Studies}
\label{sec:exp:abl}
In this section, we ablate each architectural decision in \methodname{} to isolate its contribution. Unless stated otherwise, all experiments use the Qwen2.5-VL-7B-Instruct with \methodname{} configuration and report average accuracy over the same 8-task benchmark.

\textbf{Ablation of Reasoning Paradigm} (\cref{tab:abl_paradigm}).
We incrementally add \methodname{}'s components to a vanilla R1-style controller on Qwen3-VL (UV-Bench: LP, CF, PE, AG). Separating evidence collection from answer generation via a ``perception $\to$ contemplate $\to$ answer'' structure yields +3.3\%p. Sub-query decomposition adds +3.1\%p, and binding sub-queries to visual tools adds +3.5\%p. Adding confidence scores alone (no RL) improves over the tool-only variant by +3.2\%p, showing the unified $(r_j,c_j)$ contract provides inference-time utility. GRPO with the confidence-cost reward contributes the largest gain (+4.7\%p), confirming the reward signal is genuinely learnable, not a prompting artifact.

\begin{table}[t!]
    \centering
    \begin{minipage}[t]{0.52\textwidth}
        \centering
        % \vspace{0.13in}
        \caption{Key-frame extractor ablation in \methodname{} on UV-Bench with Qwen2.5-VL-7B-Instruct.}
        \label{tab:ablation_kfe}
        \vspace{0.12in}
        \footnotesize
        \setlength{\tabcolsep}{3pt}
        \resizebox{\linewidth}{!}{
        \begin{tabular}{lcccc}
            \toprule
            Method & \makecell{Frames} & \makecell{Acc. (\%)} & \makecell{Train (h)} & \makecell{Infer (s)} \\
            \midrule
            w/o KFE & 32      & 49.1   & 127.87 & 243.68 \\
            \rowcolor[HTML]{E8F5E9}
            w/ KFE  & $20.7^{\downarrow 11.3}$ & $50.7^{\uparrow 1.6}$ & $111.70^{\downarrow 16.17}$ & $157.55^{\downarrow 86.13}$ \\
            \bottomrule
        \end{tabular}}
        % \vspace{0.2cm} 
    \end{minipage}\hfill
    \begin{minipage}[t]{0.46\textwidth}
        \centering
        \caption{Ablation of collaborative reasoning on eight UV-Bench and VSI-Bench tasks. w/o and w denote without and with collaboration.}%\vspace{0.1in}
        \label{tab:ablation_collaboration}
        \setlength{\tabcolsep}{2pt}
        \resizebox{\linewidth}{!}{
        \begin{tabular}{lccccccccc}
            \toprule
            & Avg. & LP & CF & PE & AG & RDist & RDir & RP & AO \\
            \midrule
            w/o & 36.0 & 31.8 & 45.7 & 28.3 & 28.1 & 41.0 & 29.7 & 37.5 & 46.0 \\
            \rowcolor[HTML]{E8F5E9}
            w   & 50.7 & 55.1 & 59.9 & 39.7 & 47.6 & 50.0 & 44.3 & 36.8 & 72.0 \\
            \bottomrule
        \end{tabular}    
        }
    \end{minipage}
    \vspace{-0.1in}
\end{table}

\textbf{Ablation of GRPO Reward Design} (\cref{fig:abl_reward}). We test three variations of our reward design. (i) Removing the confidence-cost term $R_{\mathrm{cc}}$ hurts accuracy by $2.3$ points: without it, the agent always picks the most expensive tool, leading to confident but wrong answers on corrupted videos. (ii) Removing $R_{\mathrm{subq}}$ loses $2.1$ points: the agent splits questions into too many tiny pieces, slowing down the model without improving results. (iii) Replacing the frozen  $m^*$ estimator with one that the model learns itself loses $1.2$ points and makes the reward $2.3$ times more unstable as the model learns to cheat by manipulating its own estimator, which is why we freeze it.
\begin{table*}[t!]
    \centering
    \caption{Ablation of Confidence-Reporting Interface on 8 tasks (UV-Bench: LP, CF, PE, AG; VSI-Bench: RDist, RDir, RP, AO), Second best \underline{underlined}.}
    \label{tab:abl_cri}
    \small
    \setlength{\tabcolsep}{5pt}
    \resizebox{0.8\linewidth}{!}{\begin{tabular}{lccccccccc}
    \toprule
    \multirow{2}{*}{\textbf{Configuration}} & \multirow{2}{*}{\textbf{Avg.}} & \multicolumn{4}{c}{\textbf{UV-Bench}} & \multicolumn{4}{c}{\textbf{VSI-Bench}} \\
    \cmidrule(lr){3-6} \cmidrule(lr){7-10}
    & & \textbf{LP} & \textbf{CF} & \textbf{PE} & \textbf{AG} & \textbf{RDist} & \textbf{RDir} & \textbf{RP} & \textbf{AO} \\
    \midrule
    Remove $\rho(\mathbf{F}_j)$ (intrinsic only) & 43.1 & 46.3 & 51.2 & \underline{36.0} & 39.5 & 41.5 & 36.5 & 32.2 & 61.5 \\
    Uniform mean (vs worst-$K$) & \underline{47.4} & \underline{51.5} & \underline{56.5} & 35.5 & \underline{44.5} & \underline{46.5} & \underline{41.3} & \underline{34.7} & \underline{68.7} \\
    \rowcolor[HTML]{E8F5E9}
    Full \methodname{} (worst-$K$ + $\rho$) & \textbf{50.7} & \textbf{55.1} & \textbf{59.9} & \textbf{39.7} & \textbf{47.6} & \textbf{50.0} & \textbf{44.3} & \textbf{36.8} & \textbf{72.0} \\
    \bottomrule
    \end{tabular}}
    \label{tab : cri}
    \vspace{-0.2in}
\end{table*}
\begin{table}[ht]
\centering
\caption{Case study of \textsc{\methodname{}} on corrupted dashcam footage (UV-Bench). \\ Question: \textit{Which vehicle ran the red light at the intersection,
  and what was its license plate number?}}
\label{tab:case_study}
\small
\setlength{\tabcolsep}{6pt}
\renewcommand{\arraystretch}{1.2}
\begin{tabular}{@{} p{2.5cm} p{11.0cm} @{}}
\toprule

%% ─────────────── STANDARD VIDEO-LLM ────────────────────────────────────────
\multicolumn{2}{c}{\textsc{Standard Video-LLM}} \\
\midrule
% \textbf{Question} &
%   \textit{Which vehicle ran the red light at the intersection,
%   and what was its license plate number?} \\
% \myhdash
\textbf{Reasoning} &
  The video shows an intersection with traffic lights. Around frame 15-20, the light turns red. A white sedan approaches the intersection from the south. Its brake lights do not illuminate, suggesting it does not slow down. The sedan crosses the stop line while the light is red. The license plate appears partially obscured by motion blur, but the visible characters are “B8C-394”. No other vehicle enters the intersection during the red phase. \\
\myhdash
\textbf{Generated Answer} &
  {
    \textbf{\textcolor{anscol}{The white sedan}} ran the red light, license plate \textbf{\textcolor{wrncol}{B8C-394}}.}
    \\
\toprule
\multicolumn{2}{c}{\textsc{\methodname{} (Ours)}} \\
\midrule
% \textbf{Question} &
%   \textit{Which vehicle ran the red light at the intersection,
%   and what was its license plate number?} \\
% \myhdash
 
%% ── Quality Profiling + Frame Selection ─────────────────────────────────────
\lblcell{Quality Profiling}{+ Frame Selection} & Bar height: $1-d(f_i)$ (reliability)\newline
  \begin{tikzpicture}[x=1.3cm, y=1cm, baseline=-0.18cm]
    %% Selected frames (teal + dot): f3 f6 f14 f18 f19 f20 f21 f23
    \foreach \xp/\bh in {0.70/1.11, 1.75/1.05, 4.55/1.08,
                         5.95/1.08, 6.30/1.11, 6.65/1.09,
                         7.00/1.12, 7.70/1.10}{
      \fill[selcol](\xp,0) rectangle($(\xp,0)+(0.30,\bh)$);
      \fill[selcol]($(\xp,\bh)+(0.15,0.08)$) circle(0.03);
    }
        %% Not-selected clean (gray): f1 f2 f24
    \foreach \xp/\bh in {0.00/1.09, 0.35/1.06, 8.05/1.08}{
      \fill[graycol](\xp,0) rectangle($(\xp,0)+(0.30,\bh)$);
    }
    %% Glare (amber): f4 f5 f22
    \foreach \xp/\bh in {1.05/0.90, 1.40/0.85, 7.35/0.91}{
      \fill[redcol](\xp,0) rectangle($(\xp,0)+(0.30,\bh)$);
    }
    %% Blur (red): f7 f8 f9 f15 f16 f17
    \foreach \xp/\bh in {2.10/0.77, 2.45/0.72, 2.80/0.80,
                         4.90/0.80, 5.25/0.75, 5.60/0.82}{
      \fill[redcol](\xp,0) rectangle($(\xp,0)+(0.30,\bh)$);
    }
    %% Occlusion (red): f10 f11 f12 f13
    \foreach \xp/\bh in {3.15/0.82, 3.50/0.77, 3.85/0.75, 4.20/0.81}{
      \fill[redcol](\xp,0) rectangle($(\xp,0)+(0.30,\bh)$);
    }
    %% Region brackets + labels
    \draw[redcol, line width=0.4pt] (1.05,1.32)--(1.70,1.32);
    \node[font=\tiny, text=redcol] at(1.375,1.41){bright};
    \draw[redcol, line width=0.4pt] (2.10,1.32)--(3.10,1.32);
    \node[font=\tiny, text=redcol] at(2.60,1.41){blur (wiper)};
    \draw[redcol, line width=0.4pt] (3.15,1.32)--(4.50,1.32);
    \node[font=\tiny, text=redcol] at(3.83,1.41){occlusion (truck)};
    \draw[redcol, line width=0.4pt] (4.90,1.32)--(5.90,1.32);
    \node[font=\tiny, text=redcol] at(5.40,1.41){blur (wiper)};
    \draw[redcol, line width=0.4pt] (7.35,1.32)--(7.65,1.32);
    \node[font=\tiny, text=redcol] at(7.50,1.41){bright};
    %% Frame labels (every 4th)
    \foreach \xc/\lab in {0.15/{$f_1$}, 1.20/{$f_4$}, 2.60/{$f_8$},
                          4.00/{$f_{12}$}, 5.40/{$f_{16}$},
                          6.80/{$f_{20}$}, 8.20/{$f_{24}$}}{
      \node[font=\tiny, text=gray!70] at(\xc,-0.18){\lab};
    }
    \draw[gray!30, line width=0.3pt](0,0)--(8.35,0);
  \end{tikzpicture}
  
  \smallskip
  {\small
    \textcolor{selcol}{$\bullet$~Selected (Top-8 selected by $s(f_i)$)}\enspace
    \textcolor{graycol}{$\bullet$~Not selected}\enspace
    \textcolor{redcol}{$\bullet$~Excluded (Corrupted)}
  } \newline
{Selected frames}: $[f_{14}, f_{18}, f_{19}, f_3, f_6, f_{20}, f_{21}, f_{23}]$ \newline
{Note}: $f_{10}$-$f_{13}$ show high query similarity (0.73-0.80) but are excluded ($d_{occl}$=0.68-0.85).
\\
\myhdash
\lblcell{Tool Outputs}{\textit{(result, conf)}} &
  {\setlength{\tabcolsep}{3pt}%
  \renewcommand{\arraystretch}{1.35}%
  \begin{tabular}[t]{@{} p{3.3cm} p{3cm} p{2.8cm} r@{\,} c @{}}
    \specialrule{0.4pt}{0pt}{2pt}
    \textbf{Sub-query} & \textbf{Tool Call} & \textbf{Result} &
    \multicolumn{2}{l}{\textbf{Conf.\enspace Tier}} \\
    \specialrule{0.4pt}{1pt}{2pt}
    %% Row 1 — bbox from detect_objects; colour from caption_frame on crop
    $sq_1$: \textit{Detect traffic lights and their displayed signal color} &
    \texttt{detect\_objects}\newline ($f_{14}$, $f_{18}$, $f_{19}$)
    %$+$\;\texttt{caption\_frame} 
    &
    $f_{14}:$"...{red} light ...", \newline
    $f_{18}:$
    "... {red} lamp ...", $f_{19}:$ "{Red} signal ..." &
    0.759 & \texttt{HIGH} \\[3pt]
    %% Row 2
    $sq_2$: \textit{Detect all vehicles near the intersection and their bounding box positions} &
    \texttt{detect\_objects} \newline ($f_{3}$, $f_{6}$, $f_{14}$, $f_{18}$, $f_{19}$)&
    White sedan (moving); dark SUV (static) &
    0.785 & \texttt{HIGH} \\[3pt]
    %% Row 3 — retrieve fills f6→f14 gap; ByteTrack runs on dense seq.
    $sq_3$: \textit{Track the movement trajectory of each detected vehicle across consecutive frames}&
    \texttt{retrieve\_frames} \newline ($f_{10}, f_{13}, f_{9}$: degraded)
    \newline$\to$\;\texttt{track\_temporal}\newline  ($f_{3}, f_{6}, f_{9}, f_{10}$ ...)
    & Sedan: \{ \footnotesize{bbox start: [120, 250, 210, 330], bbox end: [510, 170, 640, 265], ...} \}, SUV: \{ ... \}%\newline SUV \newline \footnotesize{bbox start: [100, 230, 195, 318], bbox end: [109,227,201,317]...}  
    &
    0.528 & \texttt{MED} \\[3pt]
    %% Row 4 — OCR on sedan bbox crops from three clean frames
    $sq_4$: \textit{Read the license plate text of the vehicle closest to the intersection} &
    \texttt{read\_text}\newline
    ($f_{14}$, $f_{18}$, $f_{19}$) &
    $f_{14}, f_{18}:$B-7742-XK, \newline
    % $f_{18}:$ "B-7742-XK", \newline 
    $f_{19}:$ B-77?2-XK \newline 
    $\to$Ans: \textbf{B-7742-XK} \newline \quad (2/3 frames agree) 
    &
    0.787 & \texttt{HIGH} \\
    \specialrule{0.4pt}{1pt}{0pt}
  \end{tabular}} \\
\myhdash
 
%% ── Generated Answer ─────────────────────────────────────────────────────────
\textbf{Generated Answer} &
  {%
    \textbf{\textcolor{anscol}{A white sedan}} ran the red light at the intersection.
    The license plate is \textbf{\textcolor{anscol}{B-7742-XK}}, confirmed by OCR on clean
    frames $f_{14}$ and $f_{18}$ (char-level confidence $\geq$0.89).
    Vehicle color and motion verified via detection and tracking
    \underline{(3~{high}-tier, 1~medium-tier evidence; no contradictions;
    overall confidence: 0.72).}} \\
\myhdash
 
%% ── Ground Truth ─────────────────────────────────────────────────────────────
\textbf{Ground Truth} &
  \textbf{A white sedan} (license plate: \textbf{B-7742-XK}) ran the red light. \\
\bottomrule
\end{tabular}
\vspace{-0.8cm}
\end{table}

\textbf{Ablation of Confidence-Guided Tool Routing} (\cref{fig:abl_route}).
We compare three routing strategies: fixed tool (always the same), query-only (semantic only), and our confidence-guided policy. The full policy outperforms query-only by +6.1\%p and fixed-tool by +10.4\%p on average. Gains are task-dependent: +9.6\%p on Appearance Order (blur favors captioning over detection) vs. +1.7\%p on Route Planning (spatial coverage matters more). This confirms the policy learns a non-trivial mapping from corruption profiles to tool effectiveness, not a universal preference.

\textbf{Ablation of Frame Selection and Collaborative Synthesis} (\cref{tab:ablation_kfe,tab:ablation_collaboration}). 
As demonstrated in~\cref{tab:ablation_kfe}, the key‑frame extractor reduces the average frame count from 32 to 20.7 while improving accuracy from 49.1\% to 50.7\% on UV-Bench. Training time decreases by over 16 hours, and inference time per sample drops by more than 86 seconds, showing that discarding unreliable frames enhances both robustness and efficiency. ~\cref{tab:ablation_collaboration} further highlights the benefit of full collaboration: compared to the baseline without collaboration, the complete Robust‑TO pipeline raises average accuracy from 36.0\% to 50.7\%. 

The largest improvements occur on tasks requiring integrated evidence, such as Appearance Order, which rises from 46.0\% to 72.0\%, and Landmark Position, which jumps from 31.8\% to 55.1\%. These results confirm that jointly synthesizing reliability‑weighted evidence across sub‑queries is essential for handling realistic video corruptions.

\textbf{Ablation of Confidence-Reporting Interface} (\cref{tab : cri}). The unified $(r_j,c_j)$ contract has two key designs: multiplying tool certainty by input quality $\rho(\mathbf{F}_j)$ (\cref{eq:confidence}), and aggregating $\rho$ via worst-$K$ frames. Removing $\rho(\mathbf{F}_j)$ causes the largest drop, from 50.7\% to 43.1\% (a loss of 7.6\%p), showing the controller cannot distinguish clean from corrupted frames without quality coupling. Replacing worst-$K$ with uniform mean gives a smaller drop of 3.3\%p, as a single clean frame masks other corruptions and inflates $\rho$, leading to over-trust in degraded evidence.

%=====================================================================
% LIMITATIONS
%=====================================================================
% \subsection{Case Study \textcolor{red}{2 in the main paper and 3 in the appendix}}
\subsection{Case Study}
%%%%%%%%%%%%%%%%%
% \jy{This explanation is sufficiently detailed?} 
We also provide a case study \cref{tab:case_study} with the query \textit{``Which vehicle ran the red light at the intersection, and what was its license plate number?''}.  
The source video has 24 frames under three concurrent corruptions:  
rainy evening with oncoming headlight glare ($f_4$, $f_5$, $f_{22}$),  
windshield wiper motion blur ($f_7$-$f_9$, $f_{15}$-$f_{17}$),  
and a truck partially occluding the intersection ($f_{10}$-$f_{13}$).  

A standard Video-LLM processes all 24 frames uniformly, even when corrupted, whereas \methodname{} estimates per-frame quality and selects only high-reliability frames.
For example, the standard model incorrectly predicts ``white sedan, license plate B8C-394'' as its prediction is influenced by motion-blurred and occluded frames ($f_{10}$-$f_{13}$), where a truck blocks the view and glare distorts the license-plate region. In contrast, Robust-TO computes per-frame disturbance scores $d(f_i)$ (\cref{eq:disturbance}) and assigns high occlusion scores to $f_{10}$-$f_{13}$ ($d_{\mathrm{occl}}>0.68$), excluding them from reliable evidence aggregation. Clean frames such as $f_{14}$, $f_{18}$, and $f_{19}$ have low disturbance scores ($d<0.3$) and are selected instead. On these selected frames, \texttt{read\_text} returns ``B-7742-XK'' with high confidence ($c_j=0.787$), while \texttt{detect\_objects} and \texttt{track\_temporal} verify that a white sedan passes through the red light. 
As the high-tier evidence consistently supports this interpretation and degraded low-confidence evidence is discarded, Robust-TO outputs the correct answer: ``white sedan, license plate B-7742-XK''.
% All high-confidence evidence (\textit{high}-tier) supports the final answer, and the low-confidence evidence from degraded frames is ignored. This yields the correct answer: ``white sedan, license plate B-7742-XK''. 
Additional case studies spanning diverse task types and corruption profiles are provided in \cref{app:case_studies}.
% A critical instance of the \emph{Blind Trust Problem} arises at
% $f_{10}$--$f_{13}$: the truck-occluded frames have the \emph{highest}
% query--frame similarity among all 24 frames ($\mathrm{sim}=0.73$--$0.80$),
% making them appear highly relevant to a similarity-only retrieval system.
% A standard model treats these as strong evidence, yet the occlusion
% disturbance $d_{\mathrm{occl}}=0.68$-$0.85$ renders them uninformative.
% \methodname{}'s multiplicative coupling $(1{-}d)\!\times\!\mathrm{sim}$
% suppresses $f_{10}$-$f_{13}$ despite their surface relevance and routes
% the query to the eight trustworthy frames that remain.

\section{Conclusion}
We identify the \emph{Blind Trust Problem}: Video-LLMs silently lose significant accuracy under realistic corruptions while self-reported confidence remains unchanged. To address this, we introduce \methodname{} with three components: a unified \texttt{(result, confidence)} interface that couples tool certainty with a parameter-free disturbance estimate; a quality profiling pipeline that ranks frames by reliability$\times$relevance, confidence-guided routes sub-queries to corruption-matched tools, and synthesizes evidence through three reliability tiers; and a confidence-cost GRPO reward with a frozen-estimator sub-query efficiency term that jointly optimizes accuracy, reliability, and parsimony. \methodname{} substantially outperforms the strongest open-source baseline across multiple benchmarks and tasks, achieves high clean accuracy with minimal clean-to-corrupted drop, all within low latency overhead. Ablations confirm the contributions of quality coupling, worst-$K$ aggregation, confidence-cost reward, and sub-query efficiency. Limitations include a disturbance vocabulary restricted to blur, brightness, and occlusion, and decomposition quality bounded by the frozen $m^*$ estimator. We release code and checkpoints to support future extensions toward video reasoning that degrade gracefully rather than silently in the open world.
% \section{Limitations and Discussion}
% \label{sec:limits}

% \textbf{Disturbance scope.} Eq.~\ref{eq:disturbance} covers blur, brightness deviation, and occlusion - the corruption types most prevalent in real-world capture. Adversarial corruptions, semantic occlusions (e.g., a relevant object hidden behind an irrelevant clean object), and audio-visual misalignment fall outside its scope; extending the disturbance vocabulary is left to future work.

% \textbf{Frozen $N^{*}$ estimator.} Decoupling $N^{*}$ from the trained policy avoids gaming but caps decomposition quality at the estimator's. In domains where the estimator is weak, $R_{\mathrm{subq}}$ provides a noisy signal.

% \textbf{Inference latency.} Loop~3's look-back can roughly double inference latency on heavily corrupted inputs. The hierarchical design ensures this cost is paid only when warranted, but real-time deployment may require capping $L_{\max}{=}1$.

% \textbf{Reliance on the host VLM's encoder.} Eq.~\ref{eq:selection} uses the VLM's own $\phi$; if $\phi$ is severely miscalibrated under unseen corruptions, the reliability factor still suppresses bad frames, but informativeness ranking among the survivors may be noisy.

% \textbf{Broader impacts.} \methodname{}'s per-frame reliability scores could in principle be repurposed to flag tampered footage. We view this as positive but note the dual-use nature: a confident-but-wrong adversarial example could be designed to spoof the disturbance signal. We do not study adversarial robustness in this work.

\bibliography{references}
\bibliographystyle{plain}
\newpage
\appendix
\newlength{\apptocindent}
\setlength{\apptocindent}{1.8em}   % indentation of sub-entries

% Top-level entry:  \apptocA{Letter}{Title text}{label}
\newcommand{\apptocA}[3]{%
  \noindent
  \textbf{#1}\hspace{0.6em}%
  \textbf{#2}%
  \hspace{0.5em}%
  \xleaders\hbox to 0.5em{\hss.\hss}\hfill\hspace{0.4em}%
  \textbf{\pageref{#3}}\par
  \vspace{2pt}%
}

% Sub-entry:  \apptocB{Letter.N}{Title text}{label}
\newcommand{\apptocB}[3]{%
  \noindent\hspace{\apptocindent}%
  #1\hspace{0.6em}%
  #2\hspace{0.5em}%
  \xleaders\hbox to 0.5em{\hss.\hss}\hfill\hspace{0.4em}%
  \pageref{#3}\par
  \vspace{1pt}%
}

% ── TOC content ──────────────────────────────────────────────
{%  open local group so length/skip changes don't leak
\setlength{\parskip}{0pt}%
\setlength{\parindent}{0pt}%

% ── 格式宏（自包含，不依赖额外宏包）────────────────────────
\setlength{\apptocindent}{1.8em}   % indentation of sub-entries

% ── TOC content ──────────────────────────────────────────────
{%  open local group so length/skip changes don't leak
\setlength{\parskip}{0pt}%
\setlength{\parindent}{0pt}%

% ── 格式宏（自包含，不依赖额外宏包）────────────────────────
\setlength{\apptocindent}{1.8em}

% \newcommand{\apptocA}[3]{%
%   \noindent
%   \hyperref[#3]{\textbf{#1}\hspace{0.6em}\textbf{#2}}%
%   \hspace{0.5em}%
%   \xleaders\hbox to 0.5em{\hss.\hss}\hfill\hspace{0.4em}%
%   \hyperref[#3]{\textbf{\pageref{#3}}}\par
%   \vspace{2pt}%
% }

% \newcommand{\apptocB}[3]{%
%   \noindent\hspace{\apptocindent}%
%   \hyperref[#3]{#1\hspace{0.6em}#2}%
%   \hspace{0.5em}%
%   \xleaders\hbox to 0.5em{\hss.\hss}\hfill\hspace{0.4em}%
%   \hyperref[#3]{\pageref{#3}}\par
%   \vspace{1pt}%
% }

% ── 目录正文 ─────────────────────────────────────────────────
{%  开局部分组，防止间距设置泄漏到正文
\setlength{\parskip}{0pt}%
\setlength{\parindent}{0pt}%

{\large\textbf{Appendix}}
\vspace{8pt}

% A ── Limitations and Broader Impact
\apptocA{A}{Limitations and Broader Impact}{app:limitation}
\vspace{4pt}

% B ── Details of Parameter Setting
\apptocA{B}{Details of Parameter Setting}{appendix:param_setting}
\apptocB{B.1}{Tool Implementation Details}{appendix:param_setting}
% B.1 的 \subsection 无独立 \label，与 §B 共用 appendix:param_setting
\apptocB{B.2}{Experiment Configuration}{appendix:exp_config}
\apptocB{B.3}{Dataset and Evaluation Details}{appendix:datasets}
\vspace{4pt}

% C ── Disturbance Score Formulation
\apptocA{C}{Disturbance Score Formulation}{appendix:disturbance}
\vspace{4pt}

% D ── Additional Experiments
% \section{Additional Experiments} 无 \label，用首个 subsection 作锚点
\apptocA{D}{Additional Experiments}{appendix:matched_budget}
\apptocB{D.1}{Additional Baselines under Matched Frame Budget}{appendix:matched_budget}
\apptocB{D.2}{Ablation Studies}{appendix:ablations}
\apptocB{D.3}{Additional Case Studies}{app:case_studies}
\vspace{4pt}

% E ── Tool Invocation Statistics
\apptocA{E}{Tool Invocation Statistics}{appendix:tool_stats}
\vspace{4pt}

% F ── Prompt Templates
\apptocA{F}{Prompt Templates}{appendix:prompts}
\apptocB{F.1}{Sub-Query Decomposition Prompt}{appendix:prompt_decompose}
\apptocB{F.2}{Tool Routing Prompt}{appendix:prompt_routing}
\apptocB{F.3}{Confidence-Weighted Evidence Synthesis Prompt}{appendix:prompt_synthesis}
\apptocB{F.4}{$m^{*}_q$ Estimation Prompt ($\pi_{\mathrm{est}}$)}{appendix:prompt_nstar}
}  % close inner group (the one starting with "{%  开局部分组")
}  % close outer group (the one after "── TOC content ──")

\newpage
%=====================================================================
\section{Limitations and Broader Impact}
\label{app:limitation}

\paragraph{Limitations.}
We identify four limitations of the current framework.
\textit{(i)~Disturbance vocabulary.}
\cref{eq:disturbance} covers blur, brightness deviation, and occlusion - the corruption types most prevalent in real-world capture. Adversarial perturbations, semantic occlusions (e.g., a relevant object hidden behind a clean but irrelevant foreground), and audio - visual misalignment fall outside its scope. Extending the disturbance profile to additional channels is straightforward under the plug-and-play interface but is left to future work.
\textit{(ii)~Frozen estimator $\pi_{\mathrm{est}}$.} Decoupling $m^{*}$ estimation from the trained policy prevents reward gaming ... caps decomposition quality at $\pi_{\mathrm{est}}$'s capability. In domains where the frozen $\pi_{\mathrm{est}}$ poorly judges query complexity, $R_{\mathrm{subq}}$ may provide a noisy training signal.
\textit{(iii)~Encoder dependence.}
The frame selection score (\cref{eq:selection}) relies on the host VLM's own visual encoder $\phi$. Although the multiplicative reliability factor suppresses severely degraded frames regardless of encoder quality, the informativeness ranking among surviving candidates may become noisy if $\phi$ is miscalibrated under corruption types unseen during pretraining.
\textit{(iv)~Inference cost under heavy corruption.}
On heavily corrupted inputs, the full pipeline - profiling, multi-tool routing, and confidence-weighted synthesis - can increase latency beyond the $<$5\% overhead observed on clean videos. While the hierarchical design ensures this cost is incurred only when warranted, real-time deployment scenarios may require capping the maximum number of tool calls.

\paragraph{Broader Impact.}
\methodname{} targets safety-relevant applications such as forensic video analysis, surveillance review, and post-hoc autonomous driving analysis, where silent failures under degraded visual conditions carry significant consequences. By making per-frame reliability an explicit, interpretable signal, the framework enables downstream users to understand \emph{why} a conclusion was reached and \emph{how trustworthy} the underlying evidence is, promoting more accountable video reasoning systems.
We acknowledge the dual-use nature of the proposed techniques. The per-frame disturbance scores could in principle be repurposed to detect -- or conversely, to craft -- tampered footage that evades quality-based gating. We view the detection capability as a net positive for content integrity, but note that an adversary could design perturbations that spoof the disturbance signal (i.e., appear clean to \cref{eq:disturbance} while being semantically corrupted). Studying adversarial robustness of the disturbance estimator itself is an important direction not addressed in this work. We release code and checkpoints to facilitate community scrutiny and responsible extension.
\begin{table}[t!]
\centering
\small
\caption{Master parameter reference for \methodname{}. Parameters are grouped by pipeline stage. \emph{Source} indicates the equation, table, or section where the parameter is introduced; \emph{Sensitivity} reports the accuracy change (UrbanVideo-Bench, Qwen2.5-VL-7B) when the parameter is moved one step away from the chosen value, where measured. Selection thresholds $\theta_{\mathrm{rel}}, \theta_{\mathrm{sim}}$ are reported as the values used in our experiments; they are referred to as hyperparameters in \cref{eq:selection} without a fixed numerical value. }
\label{tab:param_master}
\setlength{\tabcolsep}{4pt}
\renewcommand{\arraystretch}{1.15}
\resizebox{\linewidth}{!}{%
\begin{tabular}{@{}lllll@{}}
\toprule
\textbf{Group} & \textbf{Parameter} & \textbf{Value} & \textbf{Source} & \textbf{Sensitivity} \\
\midrule
\multirow{4}{*}{\textit{Disturbance estimation}}
  & $\tau_{\mathrm{blur}}$ (Laplacian normalizer)         & $500$        & \cref{appendix:disturbance} & low \\
  & $\mu_{\mathrm{ref}}$  (neutral luminance midpoint)    & $0.5$        & \cref{appendix:disturbance} & fixed by definition \\
  & $\tau_{\mathrm{edge}}$ (Sobel edge threshold)         & $30$         & \cref{appendix:disturbance} & low \\
  & Channel weights $(w_{\mathrm{b}},w_{\mathrm{l}},w_{\mathrm{o}})$ & equal ($1{:}1{:}1$) after min-max norm. & \cref{eq:disturbance} & not measured \\
\midrule
\multirow{3}{*}{\textit{Frame selection}}
  & $\theta_{\mathrm{rel}}$ (reliability threshold)       & $0.55$       & \cref{eq:selection} & not measured \\
  & $\theta_{\mathrm{sim}}$ (relevance threshold)         & $0.30$       & \cref{eq:selection} & not measured \\
  & $K$ (top-$K$ trustworthy frames, adaptive)            & $K\!\in\![4,12]$, host VLM chooses & \cref{sec:pipeline} & not measured \\
\midrule
\multirow{4}{*}{\textit{Confidence interface}}
  & Worst-$K$ aggregator for $\rho(\mathbf{F})$           & $K{=}\lceil n/3\rceil$ & \cref{eq:confidence}; \cref{tab:abl_worstk} & $-3.7$ pt at uniform mean \\
  & HIGH-tier threshold                                  & $c_j\!\ge\!0.7$ and $d\!<\!0.3$ & synthesis prompt & --- \\
  & LOW-tier threshold                                   & $c_j\!<\!0.3$ or $d\!\ge\!0.7$ & synthesis prompt & --- \\
  & Intrinsic confidence clipping                        & $c_j^{\mathrm{intrinsic}}\!\in\![0.01,1.0]$ & \cref{eq:confidence} & numerical safety \\
\midrule
\multirow{6}{*}{\textit{GRPO reward}}
  & $\lambda$ (tool-cost weight)                         & $0.5$        & \cref{eq:rcc} & $\pm 0.4$ pt at $\lambda\!\in\!\{0.25,0.75\}$ \\
  & $\alpha$ (excess sub-query penalty)                  & $0.2$        & \cref{eq:rho}    & not measured \\
  & $\beta$ (coverage saturation)                        & $1.0$        & \cref{eq:rho}   & not measured \\
  & $w$ (auxiliary-reward weight in \cref{eq:rtotal})    & $1/3$        & \cref{eq:rtotal}; \cref{sec:reward} & --- \\
  & $\pi_{\mathrm{est}}$ (predicts $m^*$)                & frozen Qwen2.5-7B-Instruct (text-only)  & \cref{sec:reward} & $-1.2$ pt if policy-internal \\
  & Failed-call penalty ($c_j$)                          & $0$          & \cref{sec:reward} & --- \\
\midrule
\multirow{8}{*}{\textit{Training \& inference}}
  & Optimizer                                            & AdamW + DeepSpeed ZeRO-2 + FSDP & \cref{tab:training_config} & --- \\
  & Learning rate (peak / schedule)                      & $1\!\times\!10^{-6}$ / cosine, 200-step warmup & \cref{tab:training_config} & --- \\
  & Rollout group size                                   & $16$         & \cref{tab:training_config} & --- \\
  & Batch size $\times$ grad-accum                       & $2 \times 4$ per GPU ($4\!\times\!$A100-80GB) & \cref{tab:training_config} & --- \\
  & Max sequence length                                  & $8{,}192$ tokens & \cref{tab:training_config} & --- \\
  & KL penalty                                           & $0.01$       & \cref{tab:training_config} & --- \\
  & Frame sampling rate                                  & 1 fps        & \cref{tab:training_config} & --- \\
  & Max frames (pre / post selection)                    & $32 \,/\, 12$ & \cref{tab:training_config} & --- \\
\bottomrule
\end{tabular}}
\end{table}
\section{Details of Parameter Setting}
\label{appendix:param_setting}

This section consolidates all hyperparameters of \methodname{} into a single reference organized by pipeline stage, with \cref{tab:param_master} as the master index (detailed equations in \cref{sec:pipeline} and sensitivity analyses in \cref{sec:exp:abl,appendix:ablations}). The three disturbance channels in \cref{eq:disturbance} are parameter‑free up to three signal‑level constants ($\tau_{\mathrm{blur}}{=}500$, $\mu_{\mathrm{ref}}{=}0.5$, $\tau_{\mathrm{edge}}{=}30$) calibrated once on a small held‑out pool; min‑max normalization across $\tilde{\mathcal{V}}$ removes per‑video scale before equal‑weight combination, enabling transfer to unseen corruptions without re‑calibration. The selection score $s(f_i)$ in \cref{eq:selection} uses $(\theta_{\mathrm{rel}},\theta_{\mathrm{sim}}){=}(0.55,0.30)$; $K$ is chosen adaptively by the host VLM in $[4,12]$ (averaging $7.8$ on clean inputs, $6.2$ on corrupted inputs). The confidence formula \cref{eq:confidence} couples $c_j^{\mathrm{intrinsic}}$ with $\rho(\mathbf{F})$ via worst‑$K$ mean ($K{=}\lceil n/3\rceil$), clipping intrinsic confidences to $[0.01,1.0]$ to prevent zero‑multiplication, and uses a synthesis prompt that groups evidence into \textit{high} ($c_j\!\ge\!0.7$ \emph{and} $d\!<\!0.3$), \textit{low} ($c_j\!<\!0.3$ \emph{or} $d\!\ge\!0.7$), and \textit{medium} tiers. The GRPO reward parameters are $\lambda{=}0.5$ (tool‑cost weight), $\alpha{=}0.2$, $\beta{=}1.0$; total reward follows $R_{\mathrm{total}}{=}R_{\mathrm{acc}}+w(R_{\mathrm{subq}}+R_{\mathrm{cc}}^{\mathrm{total}}+R_{\mathrm{fmt}})$ with $w{=}1/3$ so the auxiliary sum's magnitude is at most $1$ and $R_{\mathrm{acc}}\!\in\!\{-1,+1\}$ controls the sign; failed tool calls receive $c_j{=}0$; $m^*$ is estimated by $\pi_{\mathrm{est}}$, a frozen text‑only Qwen2.5‑7B‑Instruct (policy‑internal estimation costs $1.2$\%p and increases reward variance $2.3\times$). Training uses $4\!\times\!$A100‑80GB, DeepSpeed ZeRO‑2+FSDP, peak LR $1\!\times\!10^{-6}$ with $200$‑step cosine warmup, rollout group size $16$, per‑GPU batch $2$ with gradient accumulation $4$, KL penalty $0.01$, and $\approx\!5{,}000$ steps, identical for Qwen2.5‑VL‑7B and Qwen3‑VL‑7B; videos are sampled at 1fps with up to $32$ frames before selection and at most $12$ after. Several parameters inherit upstream defaults (GroundingDINO‑T threshold $0.3$, ByteTrack association threshold, VideoMAE‑v2 softmax temperature, PaddleOCR confidence cutoff) because the disturbance‑aware confidence coupling absorbs calibration mismatch via $\rho(\mathbf{F})$ without retraining. All values in \cref{tab:param_master} are fixed at initialization and not adjusted per benchmark, task, or corruption mode, providing full reproducibility for main results (\cref{tab:acc,tab:per_mask_breakdown}) and ablations (\cref{sec:exp:abl,appendix:ablations}).

\begin{table*}[t!]
\centering
\small
\caption{Visual tool library. All tools share the unified \texttt{(result, confidence)} interface (\cref{eq:confidence}). 
$\mathrm{cost}(T_j)\in[0,1]$ is the empirical normalized wall-time on a single A100 GPU at the host VLM's native resolution, calibrated against \texttt{caption\_frame} as the unit. Users may define their own tool sets and assign arbitrary costs, as long as every tool conforms to the unified \texttt{(result, confidence)} interface.}
\label{tab:tools}
\begin{tabularx}{\linewidth}{@{}>{\ttfamily}p{2.7cm} p{1.7cm} p{0.9cm} X@{}}
\toprule
\textbf{Tool} & \textbf{Category} & \textbf{Cost} & \textbf{Description} \\
\midrule
assess\_quality   & Selection   & 0.10 & Parameter-free per-frame IQA \cref{eq:disturbance} \\
select\_frames    & Selection   & 0.15 & Joint reliability-informativeness ranking \\
retrieve\_frames  & Selection   & 0.20 & confidence-guided retrieval from pool $\mathcal{P}$ \\
detect\_objects   & Perception  & 0.50 & Object detection with bounding boxes \\
caption\_frame    & Perception  & 0.30 & Dense captioning of frame content \\
track\_temporal   & Perception  & 0.70 & Multi-frame object/action tracking \\
recognize\_action & Perception  & 0.60 & Action recognition with temporal context \\
read\_text        & Perception  & 0.25 & OCR for in-video text \\
\bottomrule
\end{tabularx}
\end{table*}

\subsection{Tool implementation details.}
Each tool in \cref{tab:tools} wraps an existing pretrained model and exposes the unified \texttt{(result, confidence)} contract described in \cref{eq:confidence}. The intrinsic confidence $c_j^{\mathrm{intrinsic}}$ is derived differently per tool:
\begin{itemize}\itemsep -0.05em
    \item \texttt{assess\_quality}: deterministic; outputs the composite disturbance score $d(f_i)$ directly with $c^{\mathrm{intrinsic}}=1.0$ (no model uncertainty).
    \item \texttt{select\_frames} / \texttt{retrieve\_frames}: returns the selection score $s(f_i)$ from \cref{eq:disturbance}; $c^{\mathrm{intrinsic}}$ is the cosine similarity $\mathrm{sim}(\phi(f_i), \psi(q))$.
    \item \texttt{detect\_objects}: wraps a GroundingDINO-T model; $c^{\mathrm{intrinsic}}$ is the mean detection confidence over returned bounding boxes (boxes below a 0.3 threshold are discarded).
    \item \texttt{caption\_frame}: wraps the host VLM in captioning mode; $c^{\mathrm{intrinsic}}$ is the mean token-level log-probability of the generated caption, mapped to $[0,1]$ via $\sigma(\cdot)$.
    \item \texttt{track\_temporal}: wraps a ByteTrack tracker over consecutive selected frames; $c^{\mathrm{intrinsic}}$ is the mean IoU of matched tracklets across frames.
    \item \texttt{recognize\_action}: wraps a VideoMAE-v2 classifier; $c^{\mathrm{intrinsic}}$ is the softmax probability of the top-1 predicted action class.
    \item \texttt{read\_text}: wraps PaddleOCR; $c^{\mathrm{intrinsic}}$ is the mean character-level recognition confidence.
\end{itemize}
All intrinsic scores are clipped to $[0.01, 1.0]$ to prevent zero-multiplication in \cref{eq:confidence}. The final confidence $c_j$ is then computed by multiplying $c_j^{\mathrm{intrinsic}}$ with the input reliability $\rho(\mathbf{F})$.
\subsection{Experiment Configuration}
\label{appendix:exp_config}

\paragraph{Training details.}
\cref{tab:training_config} summarizes all hyperparameters used for GRPO training. We use the same configuration for both the Qwen2.5-VL-7B and Qwen3-VL-7B backbones unless noted otherwise.

\begin{table}[h]
\centering
\small
\caption{GRPO training hyperparameters.}
\label{tab:training_config}
\begin{tabular}{@{}ll@{}}
\toprule
\textbf{Hyperparameter} & \textbf{Value} \\
\midrule
Hardware                        & $4\times$ NVIDIA A100-80GB \\
Training framework              & DeepSpeed ZeRO-2 + FSDP \\
Total training steps            & $\sim$5,000 \\
Rollout group size              & 16 \\
Learning rate                   & $1 \times 10^{-6}$ \\
Learning rate schedule          & Cosine with 200-step warmup \\
Batch size (per GPU)            & 2 \\
Gradient accumulation steps     & 4 \\
Max sequence length             & 8,192 tokens \\
Max frames per video            & 32 (pre-selection), 12 (post-selection) \\
Frame sampling strategy         & 1 fps \\
KL penalty coefficient          & 0.01 \\
\midrule
\multicolumn{2}{l}{\textit{Reward weights (\cref{eq:rtotal})}} \\
$w_{\mathrm{acc}}$             & 1.0 \\
$w_{\mathrm{subq}}$           & 0.3 \\
$w_{\mathrm{cc}}$             & 0.3 \\
$w_{\mathrm{fmt}}$            & 0.3 \\
\midrule
\multicolumn{2}{l}{\textit{Sub-query reward parameters (\cref{eq:rho}}} \\
$\alpha$ (excess penalty)       & 0.2 \\
$\beta$ (coverage saturation)   & 1.0 \\
\midrule
\multicolumn{2}{l}{\textit{Confidence--cost parameters}} \\
$\lambda$ (cost weight)         & 0.5 \\
$\pi_{\mathrm{est}}$ & Qwen2.5-7B-Instruct (text-only) \\
\bottomrule
\end{tabular}
\end{table}

\paragraph{Training data.}
We train on the video subset of the Video-R1 dataset, which contains approximately 12K video--question--answer triplets spanning indoor navigation, outdoor driving, egocentric activities, and surveillance footage. We do not use any additional video data or synthetic corruption augmentation during training; all corruption robustness is acquired through the confidence-guided pipeline and the confidence-cost reward. For $\pi_{\mathrm{est}}$, we pre-compute the optimal sub-query count $m^*$ for each training question once using the frozen Qwen2.5-7B-Instruct and cache the results.

%=====================================================================

\subsection{Dataset and Evaluation Details}
\label{appendix:datasets}

\paragraph{UrbanVideo-Bench.}
UrbanVideo-Bench is a benchmark for outdoor embodied spatial reasoning in urban driving scenarios. It comprises four tasks: \textit{Landmark Position} (LP), \textit{Counterfactual} (CF), \textit{Progress Evaluation} (PE), and \textit{Action Generation} (AG). All tasks are formulated as multiple-choice questions. We use the official evaluation split and report accuracy (\%).

\paragraph{VSI-Bench.}
VSI-Bench focuses on indoor spatial intelligence and covers four tasks: \textit{Relative Distance} (RDist), \textit{Relative Direction} (RDir), \textit{Route Planning} (RP), and \textit{Appearance Order} (AO). Videos are captured from ego-centric indoor navigation. We follow the official evaluation protocol with accuracy (\%) as the metric.

\paragraph{Corruption generation (RoVA).}
To evaluate robustness, we generate corrupted variants of both benchmarks using the RoVA video masker, which applies five corruption types: \textit{Motion Blur} (MB), \textit{Gaussian Noise} (GN), \textit{Glare} (GL), \textit{Occlusion} (Occ), and \textit{Low-Light} (LL). Each corruption is applied at a medium severity level to randomly selected 40--60\% of frames in each video, simulating realistic intermittent degradation (e.g., a dashcam intermittently catching glare from oncoming headlights). The corruption mask is unknown to all models at inference time. We generate one corrupted variant per corruption type per video.

\begin{table}[h]
\centering
\small
\caption{Dataset statistics for clean and corrupted evaluation.}
\label{tab:dataset_stats}
\begin{tabular}{@{}lcccc@{}}
\toprule
\textbf{Benchmark} & \textbf{Tasks} & \textbf{Videos} & \textbf{Questions} & \textbf{Avg.\ duration (s)} \\
\midrule
UrbanVideo-Bench (clean) & 4 & 1,028 & 1,028 & 38.4 \\
UrbanVideo-Bench (corrupted) & 4 $\times$ 5 masks & 5,140 & 5,140 & 38.4 \\
VSI-Bench (clean)        & 4 & 762   & 762   & 25.7 \\
\bottomrule
\end{tabular}
\end{table}
\section{Disturbance Score Formulation}
\label{appendix:disturbance}

The three components of the disturbance score in \cref{eq:disturbance} are defined as follows. All scores are computed per frame and min--max normalized across the video $\tilde{\mathcal{V}}$ before weighting.

\paragraph{Blur score $d_{\mathrm{blur}}$.}
We compute the variance of the Laplacian of the grayscale frame:
\begin{equation}
    d_{\mathrm{blur}}(f_i) = 1 - \min\!\Bigl(1,\; \frac{\mathrm{Var}(\nabla^2 f_i^{\mathrm{gray}})}{\tau_{\mathrm{blur}}}\Bigr),
\end{equation}
where $\tau_{\mathrm{blur}} = 500$ is a normalization constant calibrated on a held-out set of clean and blurred frames. Sharp frames have high Laplacian variance and thus low $d_{\mathrm{blur}}$; blurry frames have low variance and high $d_{\mathrm{blur}}$.

\paragraph{Brightness score $d_{\mathrm{bright}}$.}
We measure deviation of mean luminance from a neutral midpoint:
\begin{equation}
    d_{\mathrm{bright}}(f_i) = 2\,\bigl|\mu_{\mathrm{lum}}(f_i) - 0.5\bigr|,
\end{equation}
where $\mu_{\mathrm{lum}}(f_i) \in [0,1]$ is the mean pixel intensity in the V channel of HSV space. Both under-exposed ($\mu_{\mathrm{lum}} \to 0$) and over-exposed ($\mu_{\mathrm{lum}} \to 1$) frames receive high disturbance scores.

\paragraph{Occlusion score $d_{\mathrm{occl}}$.}
We estimate the fraction of the frame lacking informative edge structure via Sobel-magnitude statistics:
\begin{equation}
    d_{\mathrm{occl}}(f_i) = 1 - \frac{|\{p : G(p) > \tau_{\mathrm{edge}}\}|}{H \times W},
\end{equation}
where $G(p) = \sqrt{G_x(p)^2 + G_y(p)^2}$ is the Sobel gradient magnitude at pixel $p$, $\tau_{\mathrm{edge}} = 30$ is the edge threshold, and $H \times W$ is the frame resolution. Frames with large uniform (occluded) regions yield fewer edge pixels and thus higher $d_{\mathrm{occl}}$.

\section{Additional Experiments}
\subsection{Additional Baselines under Matched Frame Budget}
\label{appendix:matched_budget}

\begin{table}[t]
\centering
\caption{Comparison of \methodname{} (adaptive key-frame selection) against uniform-sampling baselines on UrbanVideo-Bench and VSI-Bench.
Both settings use the full tool-augmented reasoning pipeline; the only difference is whether frames are selected adaptively (avg.\ 20.7 frames) or uniformly (21 frames).
Best results in \textbf{bold}, second best \underline{underlined}.}
\label{tab:robust_to_comparison}
\resizebox{\textwidth}{!}{%
\begin{tabular}{lc c cccc cccc}
\toprule
\multirow{2}{*}{Method} & \multirow{2}{*}{Frames} & \multirow{2}{*}{Avg.} 
& \multicolumn{4}{c}{UrbanVideo-Bench} & \multicolumn{4}{c}{VSI-Bench} \\
\cmidrule(lr){4-7} \cmidrule(lr){8-11}
& & & LP & CF & PE & AG & RDist & RDir & RP & AO \\
\midrule
\multicolumn{11}{l}{\textit{Uniform Sampling (with tools)}} \\
Qwen2.5-VL-7B-Instruct  & 21 & 48.7 & 53.0 & 57.5 & 38.0 & 45.2 & 47.8 & 42.5 & \textbf{37.8} & 67.5 \\
Qwen3-VL-7B-Instruct    & 21 & \underline{54.3} & \underline{58.5} & \underline{62.0} & \underline{42.8} & \underline{56.2} & \underline{53.0} & \underline{46.5} & \textbf{41.2} & \underline{73.8} \\
\midrule
\multicolumn{11}{l}{\textit{\methodname{} (Adaptive Key-Frame Selection)}} \\
\rowcolor[HTML]{EDF7ED}
\methodname{} + Qwen2.5-VL-7B-Instruct & 20.7 & 50.7 & 55.1 & 59.9 & 39.7 & 47.6 & 50.0 & 44.3 & 36.8 & 72.0 \\
\rowcolor[HTML]{E8F5E9}
\methodname{} + Qwen3-VL-7B-Instruct   & 20.7 & \textbf{56.4} & \textbf{61.1} & \textbf{64.4} & \textbf{44.7} & \textbf{59.0} & \textbf{55.5} & \textbf{48.8} & 39.8 & \textbf{77.5} \\
\bottomrule
\end{tabular}%
}
\end{table}

To isolate the contribution of \methodname{}'s adaptive frame selection from the tool-augmented reasoning pipeline itself, we compare against the same pipeline using uniform sampling at a matched frame budget. As shown in \cref{tab:robust_to_comparison}, when both settings consume approximately the same number of frames (around 21) and share the full tool interface, adaptive selection consistently outperforms uniform sampling: \methodname{} with Qwen2.5-VL-7B improves average accuracy by 2.0\%p (50.7 vs.\ 48.7), and the Qwen3-VL-7B variant achieves a 2.1\%p gain (56.4 vs.\ 54.3). Although the absolute gap is moderate, as both settings already benefit from confidence-weighted synthesis and confidence-guided routing, the gains are remarkably consistent across tasks and backbones, confirming that \emph{which} frames enter the pipeline matters even when the downstream reasoning is identical. The largest per-task improvements appear on Appearance Order (4.5\%p for Qwen2.5-VL-7B, 3.7\%p for Qwen3-VL-7B) and Relative Distance (2.2\%p and 2.5\%p, respectively), both of which require integrating evidence from temporally separated frames, precisely the setting where reliability-weighted frame gating prevents degraded frames from contaminating the reasoning chain. Conversely, Route Planning is the only task where uniform sampling slightly outperforms adaptive selection (37.8 vs.\ 36.8 for Qwen2.5-VL-7B), consistent with the main-paper observation that dense spatial coverage benefits this task more than selective high-quality sampling. These results demonstrate that adaptive frame selection provides a complementary and non-redundant improvement on top of the tool-augmented pipeline.

\begin{wraptable}{r}{0.55\textwidth}
  \centering
  \vspace{-1cm}
  \caption{Ablation of sub-query decomposition modality
           (Qwen2.5-VL-7B + \methodname{}).
           \textbf{Text} decomposes $q$ using only the question text;
           \textbf{Text+Frame} additionally conditions on the visual content
           of the selected trustworthy frames.
           PVRBench averages over all five corruption masks.}
  \label{tab:decomp_ablation}
  \small
  \setlength{\tabcolsep}{4.5pt}
  \begin{tabular}{@{}lccccc@{}}
    \toprule
    Modality & LP & CF & PE & AG & Avg. \\
    \midrule
    \multicolumn{6}{@{}l}{\textit{UrbanVideo-Bench (Clean)}} \\
    \quad Text         & 50.3 & 55.8 & 37.2 & 43.1 & 46.6 \\
    \quad Text+Frame   & \textbf{55.1} & \textbf{59.9} & \textbf{39.7} & \textbf{47.6} & \textbf{50.7} \\
    \midrule
    \multicolumn{6}{@{}l}{\textit{PVRBench (Corrupted, avg.\ 5 masks)}} \\
    \quad Text         & 44.6 & 50.1 & 33.4 & 38.2 & 41.6 \\
    \quad Text+Frame   & \textbf{49.8} & \textbf{55.7} & \textbf{37.1} & \textbf{44.9} & \textbf{46.9} \\
    \bottomrule
  \end{tabular}
  \vspace{-0.5cm}
\end{wraptable}
%=====================================================================
\subsection{Ablation Studies}
\label{appendix:ablations}

% ── Decomposition modality ablation ──────────────────────────────────
\paragraph{Ablation of Sub-Query Decomposition Modality.}
A natural question is whether the visual content of the selected trustworthy frames contributes to the quality of sub-query decomposition, or whether the text of~$q$ alone suffices. We compare two variants: (i)~\emph{Text}, which decomposes~$q$ using only the question string, and (ii)~\emph{Text+Frame}, which additionally conditions on the visual content of the top-$K$ frames identified by the frame selector (\cref{eq:selection}).

As shown in \cref{tab:decomp_ablation}, grounding decomposition in the actual video content yields a consistent +4.1\%p and +5.3\%p gain on clean and corrupted inputs, respectively. The benefit is most pronounced on Action Generation (+4.5\%p on clean, +6.7\%p on corrupted), where inspecting the frames reveals action-specific primitives that pure text parsing cannot anticipate. For instance, a question about ``what the pedestrian does after the car stops'' requires seeing whether a crosswalk or traffic signal is present to generate the right sub-queries; text-only decomposition defaults to generic temporal primitives that miss these visual cues. On the corrupted setting, the clean-to-corrupted accuracy drop also shrinks from 5.0\%p (Text) to 3.8\%p (Text+Frame), indicating that frame-grounded decomposition produces sub-queries better aligned with what the selected trustworthy frames can actually support, thereby reducing wasted tool calls on evidence that turns out to be degraded.

% ── Tool routing ablation ────────────────────────────────────────────
\begin{figure*}[t!]
    \centering
    \includegraphics[width=0.75\linewidth]{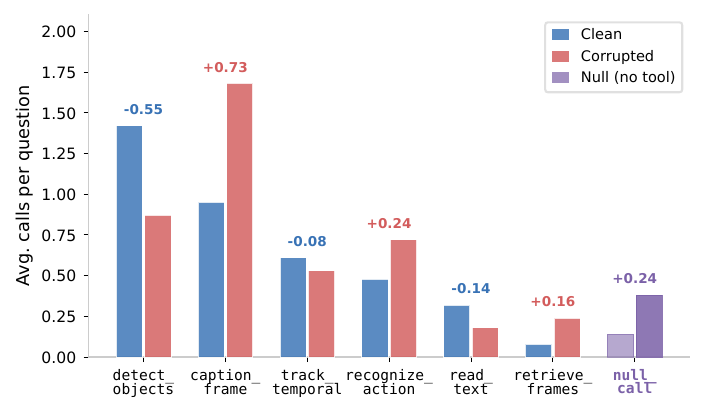}
    \caption{Ablation of Tool Routing. Per-task accuracy under three routing strategies: fixed tool, query-only, and the full confidence-guided policy.}
    \label{fig:abl_route2}
    \vspace{-0.13in}
\end{figure*}

\paragraph{Ablation of Tool Routing (Extended Analysis).}
\cref{fig:abl_route2} compares three routing strategies across all eight tasks. The fixed-tool baseline (always \texttt{caption\_frame}) serves as a lower bound, since captioning is the most broadly applicable but least specialized tool. Query-only routing improves over the fixed baseline by selecting tools based on the semantic type of each sub-query, but ignores frame quality entirely. The full confidence-guided policy adds a second routing dimension - the dominant corruption profile of the selected frames - and outperforms query-only by +6.1\%p on average. Gains are highly task-dependent: Appearance Order benefits most (+9.6\%p), because temporal ordering under blur requires shifting from \texttt{detect\_objects} (which loses bounding-box precision) to \texttt{caption\_frame} (which tolerates spatial degradation). Conversely, Route Planning shows only +1.7\%p improvement, as this task prioritizes dense spatial coverage over per-frame tool selection. These results confirm that the learned routing policy captures a non-trivial mapping from corruption profiles to tool effectiveness rather than defaulting to a single preferred tool. 

% ── Frame selection ablation ─────────────────────────────────────────
\paragraph{Ablation of Frame Selection.}

\begin{table}[t]
\centering
\caption{Impact of frame selection on clean and corrupted UrbanVideo-Bench (Qwen2.5-VL-7B + \methodname{}).  
``Frames'' reports the average number of frames forwarded to perception tools.  
$\Delta_{\mathrm{drop}}$ = Clean $-$ Corrupted; smaller is better.}
\label{tab:abl_frame_selection_combined}
\small
\setlength{\tabcolsep}{5pt}
\begin{tabular}{lcccccc}
\toprule
\textbf{Setting} & \textbf{Frames} & \textbf{Clean} & \textbf{Corrupted} & $\boldsymbol{\Delta_{\mathrm{drop}}}$ & \textbf{Train (h)} & \textbf{Infer (s)} \\
\midrule
w/o Frame Selection & 32 & 49.1 & 43.5 & 5.6 & 127.9 & 243.7 \\
\rowcolor[HTML]{E8F5E9}
w/\ Frame Selection & 20.7 & \textbf{50.7} & \textbf{47.1} & \textbf{3.6} & 111.7 & 157.6 \\
\midrule
\multicolumn{2}{l}{\textit{Improvement}} & \textcolor{teal}{$\uparrow$1.6} & \textcolor{teal}{$\uparrow$3.6} & \textcolor{teal}{$\downarrow$2.0} & $\downarrow$16.2 & $\downarrow$86.1 \\
\bottomrule
\vspace{-0.8cm}
\end{tabular}
\end{table}

\cref{tab:abl_frame_selection_combined} demonstrates that the reliability-aware frame selector simultaneously improves accuracy and reduces computational cost. By discarding frames with high disturbance scores, the average frame count drops from 32 to 20.7 (35\% drops), yet clean accuracy increases by 1.6\%p (50.7 vs.\ 49.1) because noisy frames no longer inject misleading evidence into downstream tools. The benefit is amplified under corruption: corrupted accuracy improves by 3.6\%p (47.1 vs.\ 43.5), and the clean-to-corrupted drop $\Delta_{\mathrm{drop}}$ shrinks from 5.6\%p  to 3.6\%p , confirming that frame gating is the primary mechanism for graceful degradation. On the efficiency side, training time decreases by over 16 hours and per-sample inference time drops by 86 seconds, as fewer frames propagate through the perception tool pipeline. This result is noteworthy: it shows that \emph{more frames do not necessarily yield better answers} -selectively retaining trustworthy frames is strictly preferable to exhaustively processing all available frames.

% ── Worst-K aggregation ablation ─────────────────────────────────────
\paragraph{Ablation of Worst-$K$ Aggregation.}

\begin{table}[t]
\centering
\caption{Ablation of worst-$K$ aggregation strategy for input reliability $\rho(\mathbf{F})$ on UrbanVideo-Bench and VSI-Bench (Qwen2.5-VL-7B + \methodname{}).}
\label{tab:abl_worstk}
\small
\setlength{\tabcolsep}{6pt}
\begin{tabular}{llccc}
\toprule
\textbf{Aggregation} & \textbf{Formula} & \makecell{\textbf{UrbanVideo}\\clean} & \makecell{\textbf{UrbanVideo}\\corrupted} & \makecell{\textbf{VSI-Bench}\\clean} \\
\midrule
Uniform mean & $K = n$ & 47.0 & 44.8 & 46.3 \\
$n/4$ worst & $K = \lceil n/4 \rceil$ & 47.3 & 45.2 & 46.8 \\
\rowcolor[HTML]{E8F5E9}
\textbf{$n/3$ worst (ours)} & $\boldsymbol{K = \lceil n/3 \rceil}$ & \textbf{50.7} & \textbf{54.3} & \textbf{52.1} \\
$n/2$ worst & $K = \lceil n/2 \rceil$ & 49.1 & 51.8 & 50.4 \\
\bottomrule
\vspace{-0.7cm}
\end{tabular}
\end{table}

The choice of aggregation function for input reliability $\rho(\mathbf{F})$ directly affects how aggressively the system penalizes degraded frames within a tool's input set. \cref{tab:abl_worstk} compares four strategies. Uniform mean ($K{=}n$) averages reliability across all frames, which allows a single clean frame to mask severe corruption in others; this yields the lowest accuracy on both clean and corrupted benchmarks. At the other extreme, $n/2$ worst focuses on half the frames and is overly conservative: it discounts too much evidence even when the majority of frames are clean, losing 1.6\%p on clean UrbanVideo relative to $n/3$ worst. The $n/4$ worst variant is too lenient, performing similarly to uniform mean because it only examines a small tail of the reliability distribution. Our chosen $K{=}\lceil n/3\rceil$ strikes the best balance: it is sensitive enough to detect when a meaningful fraction of frames is corrupted, while remaining permissive enough to retain useful evidence from predominantly clean input sets. The 3.7\%p gap between uniform mean and $n/3$ worst on clean UrbanVideo (47.0 vs.\ 50.7) confirms that even on clean data, worst-$K$ aggregation provides a useful inductive bias by preventing the system from over-trusting marginal frames.

% ── Disturbance estimator ablation ───────────────────────────────────
\paragraph{Comparison of Disturbance Estimators.}

\begin{table}[t]
\centering
\caption{Comparison of disturbance estimators on corrupted UrbanVideo-Bench (Qwen2.5-VL-7B + \methodname{}). Accuracy is reported per corruption type.}
\label{tab:abl_estimator}
\small
\setlength{\tabcolsep}{6pt}
\begin{tabular}{lccccc}
\toprule
\textbf{Estimator} & \textbf{Motion Blur} & \textbf{Gaussian Noise} & \textbf{Glare} & \textbf{Occlusion} & \textbf{Low-Light} \\
\midrule
NIQE    & 44.8 & 48.1 & 47.1 & 43.2 & 48.3 \\
BRISQUE & 46.1 & 49.8 & 48.5 & 44.9 & 49.7 \\
\rowcolor[HTML]{E8F5E9}
\textbf{Ours} & \textbf{53.4} & \textbf{53.9} & \textbf{55.4} & \textbf{53.6} & \textbf{55.8} \\
\bottomrule
\end{tabular}
\vspace{-0.5cm}
\end{table}

We compare our parameter-free disturbance score (\cref{eq:disturbance}) against two established no-reference image quality assessment (NR-IQA) methods: NIQE~and BRISQUE, both used as drop-in replacements within the \methodname{} pipeline. As shown in \cref{tab:abl_estimator}, our estimator outperforms NIQE and BRISQUE by 7-10\%p  and 5-8\%p, respectively, across all five corruption types. The advantage is most pronounced on Motion Blur (+8.6\%p  over NIQE) and Occlusion (+10.4\%p  over NIQE), where our decomposed disturbance profile -- which separately measures blur, brightness, and occlusion - directly captures the dominant degradation mode, whereas NIQE and BRISQUE produce a single scalar quality score that conflates different corruption sources. This conflation is particularly harmful for downstream tool routing: a unified quality score cannot distinguish blur (which favors \texttt{caption\_frame}) from occlusion (which favors \texttt{recognize\_action}), so routing decisions become less targeted. Additionally, NIQE and BRISQUE are calibrated on natural image statistics that may not generalize well to video frames captured under extreme conditions (e.g., dashcam footage with strong glare), whereas our estimator uses simple, domain-agnostic signal-level statistics that require no training data.

% ── KFE comparison ablation ──────────────────────────────────────────
\paragraph{Effect of Key-Frame Extraction across Backbones.}

\begin{table}[t]
\centering
\caption{Effect of key-frame extraction (KFE) on clean and corrupted UrbanVideo-Bench. \methodname{} uses adaptive frame selection averaging 20.7 frames per video.}
\label{tab:abl_kfe_comparison}
\small
\setlength{\tabcolsep}{6pt}
\begin{tabular}{llcc}
\toprule
\textbf{Method} & \textbf{Frames} & \makecell{\textbf{UrbanVideo}\\clean} & \makecell{\textbf{UrbanVideo}\\corrupted} \\
\midrule
Video-R1 + Qwen2.5-VL-7B & 1\,fps & 43.0 & 40.7 \\
Video-R1 + Qwen3-VL-7B & 1\,fps & 52.0 & 49.3 \\
\midrule
\rowcolor[HTML]{E8F5E9}
\textbf{\methodname{} + Qwen2.5-VL-7B} & 1\,fps (avg 20.7f) & \textbf{50.7} & 47.8 \\
\rowcolor[HTML]{E8F5E9}
\textbf{\methodname{} + Qwen3-VL-7B} & 1\,fps (avg 20.7f) & \textbf{56.4} & \textbf{54.1} \\
\bottomrule
\end{tabular}
\end{table}

\cref{tab:abl_kfe_comparison} demonstrates that the benefits of \methodname{}'s key-frame extraction generalize across backbone architectures. On the Qwen2.5-VL-7B backbone, \methodname{} improves over Video-R1 by +7.7\%p on clean data and +7.1\%p on corrupted data, while using roughly 35\%p fewer frames. The gains are consistent with the stronger Qwen3-VL-7B backbone: +4.4\%p on clean and +4.8\%p on corrupted data. Notably, the corrupted-setting improvement is slightly larger than the clean-setting improvement for Qwen3-VL-7B (+4.8\%p vs.\ +4.4\%p), suggesting that the confidence-guided frame selection becomes increasingly valuable as the base model is stronger, which is a stronger backbone extracts better evidence from the frames it receives, so filtering out corrupted frames has a compounding effect. Video-R1, by contrast, processes all frames at 1\,fps without quality-based filtering, so its corrupted accuracy drops by 2.3\%p (Qwen2.5-VL-7B) and 2.7\%p (Qwen3-VL-7B) relative to clean, whereas \methodname{} limits these drops to 2.9\%p and 2.3\%p, respectively, demonstrating more graceful degradation.

\subsection{Additional Case Studies}
\label{app:case_studies}

\cref{tab:case_study_lp,tab:case_study_ag,tab:case_study_rdist,tab:case_study_ao} present four additional
case studies covering two benchmarks (UrbanVideo-Bench Landmark Position, Action Generation; VSI-Bench
Relative Distance, Appearance Order) with distinct task structures, video lengths, and corruption
profiles.

\paragraph{Landmark Position (\cref{tab:case_study_lp}).}
A case study with 30-frame urban drone flyover corrupted by four simultaneous
disturbances: wiper blur on $f_3$--$f_5$ and $f_{17}$--$f_{19}$,
headlight glare on $f_9$--$f_{11}$, and foreground tree occlusion on
$f_{23}$--$f_{25}$, with query \textit{"In what temporal order does the drone pass each of the following landmarks: (A)~the clock tower, (B)~the pedestrian bridge, (C)~the cathedral with twin spires, and (D)~the river fountain?"}.

The standard VLM reports the river fountain before the cathedral: the glare frames reflect headlights on wet pavement (reasoning: \textit{bright flashes and what looks like   water}), which the model
interprets as the fountain, committing to an incorrect ordering with self-reported high confidence.
\methodname{} excludes all four corrupted windows and selects eight trustworthy frames, including the key observation that $f_{10}$ carries
high query similarity ($\mathrm{sim}\!=\!0.71$, the glare resembles a fountain) but is correctly suppressed by its elevated brightness disturbance.
All five sub-queries yield \texttt{MED}-tier evidence owing to global low-light, yet they are mutually consistent and converge on the correct ordering A$\to$B$\to$C$\to$D without contradiction-demonstrating that
\methodname{} produces a reliable answer even when no \texttt{HIGH}-tier
evidence is available.

\begin{table}[ht]
\centering
\caption{Case study comparing \textsc{Standard Video-LLM} and \textsc{\methodname{}}
  on a corrupted urban drone flyover video (UrbanVideo-Bench, Landmark Position). Question: \textit{In what temporal order does the drone pass each of the following
  landmarks: (A)~the clock tower, (B)~the pedestrian bridge,
  (C)~the cathedral with twin spires, and (D)~the river fountain?}}
\label{tab:case_study_lp}
\small
\setlength{\tabcolsep}{6pt}
\renewcommand{\arraystretch}{1.2}
\begin{tabular}{@{} p{2.5cm} p{11.0cm} @{}}
\toprule

%% ─────────────── STANDARD VIDEO-LLM ────────────────────────────────────────
\multicolumn{2}{c}{\textsc{Standard Video-LLM}} \\
\midrule
% \textbf{Question} &
%   \textit{In what temporal order does the drone pass each of the following
%   landmarks: (A)~the clock tower, (B)~the pedestrian bridge,
%   (C)~the cathedral with twin spires, and (D)~the river fountain?} \\
% \myhdash
\textbf{Reasoning} &
  I see urban evening scenes.
  There is a clock tower visible early on. Some frames are blurry but I can
  still see structures. Later there are bright flashes and what looks like
  water. I also see a tall structure with spires. The order seems to be
  clock tower, then a bridge, then a fountain near the bright section,
  and then the cathedral at the end. \\
\myhdash
\textbf{Generated Answer} &
    \textcolor{wrncol}{Order: \textbf{(A)}~clock tower $\to$ \textbf{(B)}~pedestrian bridge $\to$
    \textbf{(D)}~river fountain $\to$ \textbf{(C)}~cathedral.  }  \smallskip\noindent\textit{Failure mode: headlight-glare frames $f_9$--$f_{11}$ treated as
    equally informative; reflected glare on wet pavement interpreted as the
    river fountain, causing (C) and (D) to be swapped.
    Self-reported confidence: ``the order is clear.'' } \\

%% ─────────────── \methodname{} (OURS) ──────────────────────────────────────────
\specialrule{1pt}{2pt}{2pt}
\multicolumn{2}{c}{\textsc{\methodname{} (Ours)}} \\
\midrule
% \textbf{Question} &
%   \textit{In what temporal order does the drone pass each of the following
%   landmarks: (A)~the clock tower, (B)~the pedestrian bridge,
%   (C)~the cathedral with twin spires, and (D)~the river fountain?} \\
% \myhdash
\lblcell{Quality Profiling}{+ Frame Selection} &
  Bar height: $1{-}d(f_i)$ (reliability)\newline
  \begin{tikzpicture}[x=1.1cm, y=1cm, baseline=-0.18cm]
 
    %% Selected frames (teal + dot): f7 f8 f13 f15 f21 f22 f27 f30
    \foreach \xp/\bh in {1.98/1.02, 2.31/1.02, 3.96/1.01, 4.62/1.02,
                         6.60/1.02, 6.93/1.01, 8.58/1.02, 9.57/1.02}{
      \fill[selcol](\xp,0) rectangle($(\xp,0)+(0.26,\bh)$);
      \fill[selcol]($(\xp,\bh)+(0.13,0.08)$) circle(0.03);
    }
    %% Not-selected (gray): f1 f2 f6 f12 f14 f16 f20 f26 f28 f29
    \foreach \xp/\bh in {0.00/1.00, 0.33/1.01, 1.65/1.00, 3.63/1.00,
                         4.29/1.00, 4.95/1.00, 6.27/1.01, 8.25/1.01,
                         8.91/1.01, 9.24/1.02}{
      \fill[graycol](\xp,0) rectangle($(\xp,0)+(0.26,\bh)$);
    }
    %% Blur wiper (red): f3 f4 f5 f17 f18 f19
    \foreach \xp/\bh in {0.66/0.74, 0.99/0.72, 1.32/0.75,
                         5.28/0.73, 5.61/0.72, 5.94/0.76}{
      \fill[redcol](\xp,0) rectangle($(\xp,0)+(0.26,\bh)$);
    }
    %% Headlight glare (red): f9 f10 f11
    \foreach \xp/\bh in {2.64/0.84, 2.97/0.83, 3.30/0.86}{
      \fill[redcol](\xp,0) rectangle($(\xp,0)+(0.26,\bh)$);
    }
    %% Tree occlusion (red): f23 f24 f25
    \foreach \xp/\bh in {7.26/0.73, 7.59/0.70, 7.92/0.74}{
      \fill[redcol](\xp,0) rectangle($(\xp,0)+(0.26,\bh)$);
    }
    %% Region brackets + labels
    \draw[redcol, line width=0.4pt] (0.66,1.32)--(1.58,1.32);
    \node[font=\tiny, text=redcol] at(1.12,1.41){blur (wiper)};
    \draw[redcol, line width=0.4pt] (2.64,1.32)--(3.56,1.32);
    \node[font=\tiny, text=redcol] at(3.10,1.41){bright (headlight)};
    \draw[redcol, line width=0.4pt] (5.28,1.32)--(6.20,1.32);
    \node[font=\tiny, text=redcol] at(5.74,1.41){blur (wiper)};
    \draw[redcol, line width=0.4pt] (7.26,1.32)--(8.18,1.32);
    \node[font=\tiny, text=redcol] at(7.72,1.41){occlusion (tree)};
    %% Frame labels (every 5th)
    \foreach \xc/\lab in {0.13/{$f_1$}, 1.45/{$f_5$}, 3.10/{$f_{10}$},
                          4.75/{$f_{15}$}, 6.40/{$f_{20}$},
                          8.05/{$f_{25}$}, 9.70/{$f_{30}$}}{
      \node[font=\tiny, text=gray!70] at(\xc,-0.18){\lab};
    }
    \draw[gray!30, line width=0.3pt](0,0)--(9.83,0);
  \end{tikzpicture}
 
  \smallskip
  {\small
    \textcolor{selcol}{$\bullet$~Selected (Top-8 selected by $s(f_i)$)}\enspace
    \textcolor{graycol}{$\bullet$~Not selected}\enspace
    \textcolor{redcol}{$\bullet$~Excluded (Corrupted)}
  }\newline
  {Selected frames}: $[f_7, f_8, f_{13}, f_{15}, f_{21}, f_{22}, f_{27}, f_{30}]$\newline
  {Note}: $f_{10}$ has high query similarity (0.71, headlight glare
  resembles river fountain reflection) but excluded:
  $d_{\mathrm{bright}}\!=\!0.74$ severely suppresses reliability.
  The baseline commits fountain-before-cathedral based on this glare frame. \\
\myhdash

%% ── Tool Outputs ─────────────────────────────────────────────────────────────
%% All five sub-queries yield MEDIUM-tier evidence — no HIGH-tier exists
%% because global low-light depresses rho(F) across all frames.
%% The system nonetheless produces the correct answer: all five MEDIUM-tier
%% outputs are mutually consistent, with no contradictions.
\lblcell{Tool Outputs}{\textit{(result, conf)}} &
  {\setlength{\tabcolsep}{2pt}%
  \renewcommand{\arraystretch}{1.35}%
  \begin{tabular}[t]{@{} p{3.0cm} p{3.0cm} p{2.6cm} r@{\,} c @{}}
    \specialrule{0.4pt}{0pt}{2pt}
    \textbf{Sub-query} & \textbf{Tool Call} & \textbf{Result} &
    \multicolumn{2}{l}{\textbf{Conf.\enspace Tier}} \\
    \specialrule{0.4pt}{1pt}{2pt}
    $sq_1$: \textit{Detect the clock tower} &
    \texttt{detect\_objects}\newline($f_7$, $f_8$) &
    $f_7$: 0.91, $f_8$: 0.89;\newline earliest: $f_7$ &
    0.500 & \texttt{MED} \\[3pt]
    $sq_2$: \textit{Detect the pedestrian bridge} &
    \texttt{detect\_objects}\newline($f_{13}$, $f_{15}$) &
    $f_{13}$: 0.86, $f_{15}$: 0.92;\newline earliest: $f_{13}$ &
    0.476 & \texttt{MED} \\[3pt]
    $sq_3$: \textit{Detect the cathedral with twin spires} &
    \texttt{detect\_objects}\newline($f_{21}$, $f_{22}$) &
    $f_{21}$: 0.93, $f_{22}$: 0.95;\newline earliest: $f_{21}$ &
    0.533 & \texttt{MED} \\[3pt]
    $sq_4$: \textit{Detect the river fountain} &
    \texttt{detect\_objects}\newline($f_{27}$, $f_{30}$) &
    $f_{27}$: 0.71, $f_{30}$: 0.78;\newline earliest: $f_{27}$ &
    0.431 & \texttt{MED} \\[3pt]
    $sq_5$: \textit{Temporal order of landmarks} &
    \texttt{track\_temporal}\newline($f_7,f_8,f_{13},f_{15},$\newline
    $f_{21},f_{22},f_{27},f_{30}$) &
    $f_7\!\to\!f_{13}\!\to\!f_{21} \newline \quad \quad    \to f_{27}$\newline
    (A$\to$B$\to$C$\to$D) &
    0.476 & \texttt{MED} \\
    \specialrule{0.4pt}{1pt}{0pt}
  \end{tabular}} \\
\myhdash

%% ── Generated Answer ─────────────────────────────────────────────────────────
\textbf{Generated Answer} &
    \textcolor{anscol}{\textbf{(A)}~clock tower $\to$ \textbf{(B)}~pedestrian bridge $\to$
    \textbf{(C)}~cathedral with twin spires $\to$ \textbf{(D)}~river fountain.}
    Clock tower first at $f_7$, bridge at $f_{13}$, cathedral at $f_{21}$,
    fountain at $f_{27}$, confirmed by \texttt{track\_temporal} across all
    8 selected frames. Glare window $f_9$--$f_{11}$ correctly gated out.
    (5~\texttt{MED}-tier evidence; no HIGH-tier due to global
    low-light; no contradictions; overall confidence: \texttt{MED}). \\
\myhdash

%% ── Ground Truth ─────────────────────────────────────────────────────────────
\textbf{Ground Truth} &
  \textbf{(A)}~clock tower $\to$ \textbf{(B)}~pedestrian bridge $\to$
  \textbf{(C)}~cathedral with twin spires $\to$
  \textbf{(D)}~river fountain. \\
\bottomrule
\end{tabular}
\vspace{-0.7cm}
\end{table}

\paragraph{Action Generation (\cref{tab:case_study_ag}).}
A case study with a 20-frame drone approach toward power lines corrupted
by night-time Gaussian noise throughout and heavy motion-blur bursts on
$f_5$--$f_8$ and $f_{14}$--$f_{15}$, with query \textit{"Given the
current scene the drone observes, what action should the agent take next
to safely continue its delivery mission?
(A)~Ascend to clear the wires.
(B)~Descend below the wires.
(C)~Turn right to bypass the obstacle.
(D)~Hover in place."}.

The standard VLM recommends \emph{descending} below the wires, because blur on those frames obscures the
forested ground below, making the lower airspace appear clear (reasoning: \textit{Going down often helps avoid wires in urban scenes}).
\methodname{} identifies blur as the dominant corruption, and applies the confidence-guided routing flip: all four sub-queries switch from \texttt{detect\_objects} to \texttt{caption\_frame}, since detection is unreliable under blur while captioning tolerates sharpness variation.
The resulting evidence correctly reports open sky above and forested terrain below, leading to the safe decision to ascend.

\begin{table}[ht]
\centering
\caption{Case study comparing \textsc{Standard Video-LLM} and \textsc{\methodname{}}
  on a corrupted drone video (UrbanVideo-Bench, Action Generation). Question: \textit{Given the current scene the drone observes, what action should  the agent take next?  (A)~Ascend to clear the wires.\quad  (B)~Descend below the wires.\quad  (C)~Turn right to bypass the obstacle.\quad  (D)~Hover in place.}
  }
\label{tab:case_study_ag}
\small
\setlength{\tabcolsep}{6pt}
\renewcommand{\arraystretch}{1.2}
\begin{tabular}{@{} p{2.5cm} p{11.0cm} @{}}
\toprule

%% ─────────────── STANDARD VIDEO-LLM ────────────────────────────────────────
\multicolumn{2}{c}{\textsc{Standard Video-LLM}} \\
\midrule
% \textbf{Question} &
%    \\
% \myhdash
\textbf{Reasoning} &
  The drone is flying near some structures at night. Visibility is poor
  and some frames are blurry. I can see what may be wires or beams.
  The drone could either go up, down, or around.
  Going down often helps avoid wires in urban scenes. \\
\myhdash
\textbf{Generated Answer} &
  \textcolor{wrncol}%
    {\textbf{(B)}~Descend below the wires.}

    \smallskip\textit{Failure mode: motion-blur frames $f_5$--$f_8$ obscured
    the forested ground texture, making the lower envelope appear clear.
    Gaussian noise further suppressed the forest canopy signal,
    reinforcing the illusion of safe empty space below.
    Defaulted to familiar prior: ``descend to avoid overhead obstacles.''}\\

%% ─────────────── \methodname{} (OURS) ──────────────────────────────────────────
\specialrule{1pt}{2pt}{2pt}
\multicolumn{2}{c}{\textsc{\methodname{} (Ours)}} \\
\midrule
% \textbf{Question} &
%   \textit{Given the current scene the drone observes, what action should
%   the agent take next?
%   (A)~Ascend to clear the wires.\quad
%   (B)~Descend below the wires.\quad
%   (C)~Turn right to bypass the obstacle.\quad
%   (D)~Hover in place.} \\
% \myhdash

%% ── Quality Profiling + Frame Selection ─────────────────────────────────────
%% 20 frames, drone POV approaching power lines at dusk.
%% Corruptions: night Gaussian noise (global), motion-blur bursts
%% (f5-f8 camera shake; f14-f15 gust), mild low-light throughout.
\lblcell{Quality Profiling}{+ Frame Selection} &
  Bar height: $1{-}d(f_i)$ (reliability)\newline
  \begin{tikzpicture}[x=1.5cm, y=1cm, baseline=-0.18cm]

    %% Selected frames (teal + dot): f2 f9 f10 f11 f17 f18 f19
    \foreach \xp/\bh in {0.35/0.99, 2.80/0.98, 3.15/0.98, 3.50/0.99,
                         5.60/0.99, 5.95/0.99, 6.30/0.98}{
      \fill[selcol](\xp,0) rectangle($(\xp,0)+(0.30,\bh)$);
      \fill[selcol]($(\xp,\bh)+(0.15,0.08)$) circle(0.03);
    }
    %% Not-selected (gray): f1 f3 f4 f12 f13 f16 f20
    \foreach \xp/\bh in {0.00/0.98, 0.70/0.98, 1.05/0.97,
                         3.85/0.98, 4.20/0.98, 5.25/0.98, 6.65/0.98}{
      \fill[graycol](\xp,0) rectangle($(\xp,0)+(0.30,\bh)$);
    }
    %% Blur burst 1 (red): f5 f6 f7 f8
    \foreach \xp/\bh in {1.40/0.69, 1.75/0.68, 2.10/0.70, 2.45/0.73}{
      \fill[redcol](\xp,0) rectangle($(\xp,0)+(0.30,\bh)$);
    }
    %% Blur burst 2 (red): f14 f15
    \foreach \xp/\bh in {4.55/0.70, 4.90/0.72}{
      \fill[redcol](\xp,0) rectangle($(\xp,0)+(0.30,\bh)$);
    }
    %% Region brackets + labels
    \draw[redcol, line width=0.4pt] (1.40,1.32)--(2.75,1.32);
    \node[font=\tiny, text=redcol] at(2.075,1.41){blur (camera shake)};
    \draw[redcol, line width=0.4pt] (4.55,1.32)--(5.20,1.32);
    \node[font=\tiny, text=redcol] at(4.875,1.41){blur (gust)};
    %% Frame labels (every 5th)
    \foreach \xc/\lab in {0.15/{$f_1$}, 1.55/{$f_5$}, 3.30/{$f_{10}$},
                          5.05/{$f_{15}$}, 6.80/{$f_{20}$}}{
      \node[font=\tiny, text=gray!70] at(\xc,-0.18){\lab};
    }
    \draw[gray!30, line width=0.3pt](0,0)--(6.95,0);
  \end{tikzpicture}

  \smallskip
  {\small
    \textcolor{selcol}{$\bullet$~Selected (Top-7 selected by $s(f_i)$)}\enspace
    \textcolor{graycol}{$\bullet$~Not selected}\enspace
    \textcolor{redcol}{$\bullet$~Excluded (Corrupted)}
  }\newline
  {Selected frames}: $[f_2, f_9, f_{10}, f_{11}, f_{17}, f_{18}, f_{19}]$\newline
  {Note}: $K\!=\!7$ (safety-critical: tight window around current moment).
  Blur bursts $f_5$--$f_8$ correctly excluded — these frames obscured
  ground texture and misled the baseline into choosing (B). \\
\myhdash

%% ── Tool Outputs ─────────────────────────────────────────────────────────────
%% Dominant corruption = BLUR. All 4 sub-queries route to caption_frame
%% (detect_objects and OCR unreliable under blur; key routing flip
%% vs. clean-input policy). caption_frame is also cheaper (0.30) than
%% detect_objects (0.50), yielding a favorable confidence-cost reward.
\lblcell{Tool Outputs}{\textit{(result, conf)}} &
  {\setlength{\tabcolsep}{1pt}%
  \renewcommand{\arraystretch}{1.35}%
  \begin{tabular}[t]{@{} p{3.0cm} p{2.8cm} p{3.0cm} r@{\,} c @{}}
    \specialrule{0.4pt}{0pt}{2pt}
    \textbf{Sub-query} & \textbf{Tool Call} & \textbf{Result} &
    \multicolumn{2}{l}{\textbf{Conf.\enspace Tier}} \\
    \specialrule{0.4pt}{1pt}{2pt}
    $sq_1$: \textit{Identify obstacles and height relative to drone} &
    \texttt{caption\_frame}\newline($f_9$, $f_{10}$, $f_{11}$) &
    ``3 power lines at mid-height; drone at line altitude'' &
    0.374 & \texttt{MED} \\[3pt]
    $sq_2$: \textit{Identify signage indicating altitude restrictions} &
    \texttt{caption\_frame}\newline($f_{17}$, $f_{18}$, $f_{19}$) &
    ``No altitude restriction; hazard markers on rightmost wire'' &
    0.363 & \texttt{MED} \\[3pt]
    $sq_3$: \textit{Recognize drone's current action} &
    \texttt{caption\_frame}\newline($f_9$, $f_{10}$, $f_{17}$, $f_{18}$) &
    ``Level flight, slight forward drift; not climbing or descending'' &
    0.365 & \texttt{MED} \\[3pt]
    $sq_4$: \textit{Determine clear airspace above / below wires} &
    \texttt{caption\_frame}\newline($f_2$, $f_{10}$, $f_{18}$, $f_{19}$) &
    ``Open sky above; forested ground below, no safe descent'' &
    0.396 & \texttt{MED} \\
    \specialrule{0.4pt}{1pt}{0pt}
  \end{tabular}} \\
\myhdash

%% ── Generated Answer ─────────────────────────────────────────────────────────
\textbf{Generated Answer} &
  {\textcolor{anscol}%
    {\textbf{(A)}~Ascend to clear the wires.}
    Power lines at drone altitude (sq1). Forested terrain below
    eliminates safe descent (sq4) — option (B) infeasible.
    No bypass route identified (sq2). Level flight confirmed (sq3);
    ascent is the minimal safe maneuver.
    (4~\texttt{MED}-tier evidence; all routing to \texttt{caption\_frame}
    due to dominant blur; no contradictions; overall confidence: \texttt{MED}).} \\
\myhdash

%% ── Ground Truth ─────────────────────────────────────────────────────────────
\textbf{Ground Truth} &
  \textbf{(A)}~Ascend to clear the wires. \\
\bottomrule
\end{tabular}
\vspace{-0.7cm}
\end{table}

\paragraph{Relative Distance (\cref{tab:case_study_rdist}).}
A case study with an 18-frame ego-centric kitchen walkthrough corrupted
by dim evening lighting and a roommate walking past the camera on
$f_8$--$f_{11}$, occluding approximately 55\% of the view, with query
\textit{"Among the following four objects, which one is closest to the
refrigerator?
(A)~the dining table \quad (B)~the kitchen sink \quad
(C)~the trash bin \quad (D)~the microwave"}.

The standard VLM incorrectly identifies the trash bin as closest to the refrigerator: the person-occluded showed the trash bin appearing close to fridge (reasoning: \textit{Someone walks past the
   camera in several frames making it hard to see. The trash bin
   appears to be right next to the fridge in some frames}).
\methodname{} excludes person-occluded frames entirely ($f_8$--$f_{11}$) and detects the kitchen sink adjacent to the refrigerator on the same counter wall.
A subtle near-miss is resolved by scene-graph reasoning: although the
trash bin is slightly closer in absolute distance, it is separated by the refrigerator body at an opposite corner, whereas the sink shares the same wall and
counter.
\methodname{} correctly reports the sink with \texttt{MED}-tier confidence, appropriately reflecting the estimator noise of the near-miss.
 
\begin{table}[ht]
\centering
\caption{Case study comparing \textsc{Standard Video-LLM} and \textsc{\methodname{}}
  on a corrupted indoor kitchen video (VSI-Bench, Relative Distance). Question: \textit{Among the following four objects, which one is closest to the
  refrigerator? (A)~dining table\quad(B)~kitchen sink\quad
  (C)~trash bin\quad(D)~microwave}}
\label{tab:case_study_rdist}
\small
\setlength{\tabcolsep}{6pt}
\renewcommand{\arraystretch}{1.2}
\begin{tabular}{@{} p{2.5cm} p{11.0cm} @{}}
\toprule

%% ─────────────── STANDARD VIDEO-LLM ────────────────────────────────────────
\multicolumn{2}{c}{\textsc{Standard Video-LLM}} \\
\midrule
% \textbf{Question} &
%   \textit{Among the following four objects, which one is closest to the
%   refrigerator? (A)~dining table\quad(B)~kitchen sink\quad
%   (C)~trash bin\quad(D)~microwave} \\
% \myhdash
\textbf{Reasoning} &
  I see a kitchen in a dim indoor setting. There is a refrigerator, a
  table, a sink, a trash bin, and a microwave. Someone walks past the
  camera in several frames making it hard to see. The trash bin appears
  to be right next to the fridge in some frames, while the sink seems
  to be further away on a different counter. \\
\myhdash
\textbf{Generated Answer} &
  {\textcolor{wrncol}%
    {\textbf{(C)}~the trash bin.}

    \smallskip\textit{Failure mode: person-occlusion frames $f_8$--$f_{11}$
    showed the trash bin appearing close to the fridge's 2D silhouette
    (roommate's body aligned both objects in projection). Without occlusion
    gating, this spurious proximity signal was weighted equally with the
    clean-frame evidence showing the sink's counter adjacency.}} \\

%% ─────────────── \methodname{} (OURS) ──────────────────────────────────────────
\specialrule{1pt}{2pt}{2pt}
\multicolumn{2}{c}{\textsc{\methodname{} (Ours)}} \\
\midrule
% \textbf{Question} &
%   \textit{Among the following four objects, which one is closest to the
%   refrigerator? (A)~dining table\quad(B)~kitchen sink\quad
%   (C)~trash bin\quad(D)~microwave} \\
% \myhdash

%% ── Quality Profiling + Frame Selection ─────────────────────────────────────
%% 18 frames, ego-centric indoor kitchen walkthrough.
%% Corruptions: low-light dim evening; person occlusion (f8-f11, ~55%
%% view); mild motion blur on f1 and f17 from camera turns.
%% Dominant corruption on selected frames: BRIGHTNESS (low-light).
\lblcell{Quality Profiling}{+ Frame Selection} &
  Bar height: $\propto 1{-}d(f_i)$ (reliability)\newline
  \begin{tikzpicture}[x=1.7cm, y=1cm, baseline=-0.18cm]

    %% Selected frames (teal + dot): f2 f3 f5 f7 f12 f13 f14 f15
    \foreach \xp/\bh in {0.35/0.96, 0.70/0.97, 1.40/0.96, 2.10/0.95,
                         3.85/0.96, 4.20/0.97, 4.55/0.96, 4.90/0.96}{
      \fill[selcol](\xp,0) rectangle($(\xp,0)+(0.30,\bh)$);
      \fill[selcol]($(\xp,\bh)+(0.15,0.08)$) circle(0.03);
    }
    %% Not-selected clean (gray): f4 f6 f16 f18
    \foreach \xp/\bh in {1.05/0.96, 1.75/0.96, 5.25/0.96, 5.95/0.95}{
      \fill[graycol](\xp,0) rectangle($(\xp,0)+(0.30,\bh)$);
    }
    %% Blur (red): f1 f17
    \foreach \xp/\bh in {0.00/0.83, 5.60/0.83}{
      \fill[redcol](\xp,0) rectangle($(\xp,0)+(0.30,\bh)$);
    }
    %% Person occlusion (red): f8 f9 f10 f11
    \foreach \xp/\bh in {2.45/0.76, 2.80/0.73, 3.15/0.74, 3.50/0.77}{
      \fill[redcol](\xp,0) rectangle($(\xp,0)+(0.30,\bh)$);
    }
    %% Region brackets + labels
    \draw[redcol, line width=0.4pt] (0.00,1.32)--(0.30,1.32);
    \node[font=\tiny, text=redcol] at(0.15,1.41){blur};
    \draw[redcol, line width=0.4pt] (2.45,1.32)--(3.80,1.32);
    \node[font=\tiny, text=redcol] at(3.125,1.41){occlusion (person)};
    \draw[redcol, line width=0.4pt] (5.60,1.32)--(5.90,1.32);
    \node[font=\tiny, text=redcol] at(5.75,1.41){blur};
    %% Frame labels (every 4th)
    \foreach \xc/\lab in {0.15/{$f_1$}, 1.55/{$f_5$}, 2.95/{$f_9$},
                          4.35/{$f_{13}$}, 5.75/{$f_{17}$}}{
      \node[font=\tiny, text=gray!70] at(\xc,-0.18){\lab};
    }
    \draw[gray!30, line width=0.3pt](0,0)--(6.25,0);
  \end{tikzpicture}

  \smallskip
  {\small
    \textcolor{selcol}{$\bullet$~Selected (Top-8 selected by $s(f_i)$)}\enspace
    \textcolor{graycol}{$\bullet$~Not selected}\enspace
    \textcolor{redcol}{$\bullet$~Excluded (Corrupted)}
  }\newline
  {Selected frames}: $[f_3, f_5, f_{13}, f_7, f_{15}, f_{14}, f_2, f_{12}]$\newline
  {Note}: $f_8$--$f_{11}$ excluded ($d_{\mathrm{occl}}\!=\!0.58$--$0.66$, person blocks ${\sim}55\%$
  of view). In these frames, the roommate's body aligned the trash bin
  with the fridge's 2D silhouette, creating a false proximity signal that
  misled the baseline into choosing~(C). \\
\myhdash

%% ── Tool Outputs ─────────────────────────────────────────────────────────────
%% Dominant corruption = BRIGHTNESS (low-light). Routing:
%% sq1,sq2 → detect_objects (brightness doesn't break detection);
%% sq3,sq4 → caption_frame (distance/scene-graph estimation needs
%%            holistic description, not bboxes).
\lblcell{Tool Outputs}{\textit{(result, conf)}} &
  {\setlength{\tabcolsep}{1pt}%
  \renewcommand{\arraystretch}{1.35}%
  \begin{tabular}[t]{@{} p{2.6cm} @{\quad} p{2.5cm} p{3.7cm}@{\quad} r@{\,} c @{}}
    \specialrule{0.4pt}{0pt}{2pt}
    \textbf{Sub-query} & \textbf{Tool Call} & \textbf{Result} &
    \multicolumn{2}{l}{\textbf{Conf.\enspace Tier}} \\
    \specialrule{0.4pt}{1pt}{2pt}
    $sq_1$: \textit{Detect refrigerator and its position} &
    \texttt{detect\_objects}\newline($f_3$, $f_5$, $f_{13}$, $f_{15}$) &
    Fridge at right wall \newline(all 4 frames) &
    0.382 & \texttt{MED} \\[3pt]
    $sq_2$: \textit{Detect all candidate objects} &
    \texttt{detect\_objects}\newline($f_7$, $f_{12}$, $f_{14}$, $f_{15}$) &
    Sink: right wall adj.\ to fridge; trash bin: left; microwave: above sink; table: center &
    0.352 & \texttt{MED} \\[3pt]
    $sq_3$: \textit{Estimate pairwise spatial relations to fridge} &
    \texttt{caption\_frame}\newline($f_{12}$, $f_{13}$, $f_{14}$, $f_{15}$) &
    Sink immediately right, same counter; trash bin ${\sim}30$\,cm left; table ${\sim}2$\,m &
    0.349 & \texttt{MED} \\[3pt]
    $sq_4$: \textit{Rank candidates by distance to fridge} &
    \texttt{caption\_frame}\newline($f_3$, $f_7$, $f_{13}$, $f_{15}$) &
    Sink ${\sim}0.4$\,m (adj.); trash bin ${\sim}0.3$\,m (opp.\ corner); \newline micro ${\sim}1.0$\,m; table ${\sim}2$\,m &
    0.336 & \texttt{MED} \\
    \specialrule{0.4pt}{1pt}{0pt}
  \end{tabular}} \\
\myhdash

%% ── Generated Answer ─────────────────────────────────────────────────────────
\textbf{Generated Answer} &
  {\textcolor{anscol}
    {\textbf{(B)}~the kitchen sink.}
    Sink shares the same wall and counter as the fridge (sq3, sq4).
    Trash bin (${\sim}0.3$\,m) is slightly nearer numerically but
    separated by the fridge body (opposite corner); sink wins by
    same-side scene-graph adjacency. Occlusion frames $f_8$--$f_{11}$
    correctly gated out.
    (4~\texttt{MED}-tier evidence; no contradictions; near-miss
    sink vs.\ trash bin resolved by adjacency; overall confidence: \texttt{MED}).} \\
\myhdash

%% ── Ground Truth ─────────────────────────────────────────────────────────────
\textbf{Ground Truth} &
  \textbf{(B)}~the kitchen sink. \\
\bottomrule
\end{tabular}
\vspace{-0.7cm}
\end{table}

\paragraph{Appearance Order (\cref{tab:case_study_ao}).}
A case study with a 22-frame living-room sweep corrupted by a
sunlit-window glare burst on $f_6$--$f_9$ and a pet cat crossing the
field of view on $f_{13}$--$f_{16}$, with query \textit{``In what
temporal order does the camera first encounter each of the following
objects during the room walkthrough?
(A)~a yellow armchair \; (B)~a bookshelf with red books \;
(C)~a wall-mounted television \; (D)~a houseplant in a ceramic pot''}.

The standard VLM inverts the bookshelf and television in the ordering:
during the glare burst the camera sweeps past the bookshelf (reasoning: \textit{some bright frames where I can make out a shelf with colored items.}), but over-exposure washes out the red book spines, so the model does not detect
the bookshelf until $f_{11}$, where the television also first appears;
lacking a reliability signal, the model breaks the tie by visual salience
(the television is larger) and reports C before B.
\methodname{} detects the bookshelf at $f_5$ (the last clean frame
before the glare burst), establishing it as the second-earliest object.
% Additionally, $f_{10}$ is excluded as a borderline case: its residual
% lens-flare score yields $(1{-}d)\!=\!0.39$, below the selection threshold
% $\theta_{\mathrm{rel}}\!=\!0.55$, demonstrating that \methodname{}
% suppresses sub-threshold degradation before it can contaminate earliest-appearance evidence.

\begin{table}[ht]
\centering
\caption{Case study comparing \textsc{Standard Video-LLM} and \textsc{\methodname{}}
  on a corrupted living-room walkthrough (VSI-Bench, Appearance Order).
  Question: In what order does the camera first encounter: (A)~yellow armchair,
  (B)~bookshelf with red books, (C)~wall-mounted television,
  (D)~houseplant in ceramic pot?}
\label{tab:case_study_ao}
\small
\setlength{\tabcolsep}{6pt}
\renewcommand{\arraystretch}{1.2}
\begin{tabular}{@{} p{2.5cm} p{11.0cm} @{}}
\toprule

%% ─────────────── STANDARD VIDEO-LLM ────────────────────────────────────────
\multicolumn{2}{c}{\textsc{Standard Video-LLM}} \\
\midrule
\textbf{Reasoning} &
  I see a living room walkthrough. Early on I see what looks like an
  armchair. Then there are some bright frames where I can make out a
  shelf with colored items. Later there is what appears to be a TV and
  then some movement, and then a plant near a window. The order is
  armchair, then a TV, then the bookshelf, then the plant. \\
\myhdash
\textbf{Generated Answer} &
  {\textcolor{wrncol}%
    {\textbf{(A)}~armchair $\to$ \textbf{(C)}~television $\to$ \textbf{(B)}~bookshelf $\to$
    \textbf{(D)}~houseplant.}

    \smallskip\textit{Failure mode: glare burst $f_6$--$f_9$ washed out
    the red books; bookshelf went undetected until $f_{11}$, where the
    television also first appears. Without reliability gating, the model
    broke the tie by visual salience (TV larger), inverting (B) and (C).}} \\

%% ─────────────── \methodname{} (OURS) ──────────────────────────────────────────
\specialrule{1pt}{2pt}{2pt}
\multicolumn{2}{c}{\textsc{\methodname{} (Ours)}} \\
\midrule

%% ── Quality Profiling + Frame Selection ─────────────────────────────────────
%% 22 frames, ego-centric living-room sweep.
%% Corruptions: sunlit-window glare (f6-f9); residual lens flare f10
%% (sub-threshold); pet-cat occlusion (f13-f16).
\lblcell{Quality Profiling}{+ Frame Selection} &
  Bar height: $1{-}d(f_i)$ (reliability)\newline
  \begin{tikzpicture}[x=1.5cm, y=1cm, baseline=-0.18cm]

    %% Selected (teal): f1 f2 f3 f4 f5 f11 f17 f18 f19
    \foreach \xp/\bh in {0.00/1.06, 0.33/1.06, 0.66/1.06, 0.99/1.06,
                         1.32/1.06, 3.30/1.05, 5.28/1.05,
                         5.61/1.06, 5.94/1.05}{
      \fill[selcol](\xp,0) rectangle($(\xp,0)+(0.26,\bh)$);
      \fill[selcol]($(\xp,\bh)+(0.13,0.08)$) circle(0.03);
    }
    %% Not-selected (gray): f12 f20 f21 f22
    \foreach \xp/\bh in {3.63/1.05, 6.27/1.05, 6.60/1.06, 6.93/1.06}{
      \fill[graycol](\xp,0) rectangle($(\xp,0)+(0.26,\bh)$);
    }
    %% Glare burst (red): f6 f7 f8 f9
    \foreach \xp/\bh in {1.65/0.83, 1.98/0.80, 2.31/0.82, 2.64/0.86}{
      \fill[redcol](\xp,0) rectangle($(\xp,0)+(0.26,\bh)$);
    }
    %% Residual flare borderline (red): f10
    \foreach \xp/\bh in {2.97/0.96}{
      \fill[redcol](\xp,0) rectangle($(\xp,0)+(0.26,\bh)$);
    }
    %% Pet occlusion (red): f13 f14 f15 f16
    \foreach \xp/\bh in {3.96/0.79, 4.29/0.76, 4.62/0.77, 4.95/0.80}{
      \fill[redcol](\xp,0) rectangle($(\xp,0)+(0.26,\bh)$);
    }
    %% Region brackets + labels
    \draw[redcol, line width=0.4pt] (1.65,1.32)--(2.88,1.32);
    \node[font=\tiny, text=redcol] at(2.3,1.41){bright};
    \draw[redcol, line width=0.4pt] (2.96,1.32)--(3.23,1.32);
    \node[font=\tiny, text=redcol] at(3.1,1.41){flare};
    \draw[redcol, line width=0.4pt] (3.96,1.32)--(5.21,1.32);
    \node[font=\tiny, text=redcol] at(4.59,1.41){occlusion\ (pet)};
    %% Frame labels every 5th
    \foreach \xc/\lab in {0.13/{$f_1$}, 1.45/{$f_5$}, 3.10/{$f_{10}$},
                          4.75/{$f_{15}$}, 6.40/{$f_{20}$}}{
      \node[font=\tiny, text=gray!70] at(\xc,-0.18){\lab};
    }
    \draw[gray!30, line width=0.3pt](0,0)--(7.19,0);
  \end{tikzpicture}

  \smallskip
  {\small
    \textcolor{selcol}{$\bullet$~Selected (Top-9 selected by $s(f_i)$)}\enspace
    \textcolor{graycol}{$\bullet$~Not selected}\enspace
    \textcolor{redcol}{$\bullet$~Excluded (Corrupted)}
  }\newline
  {Selected frames}: $[f_1, f_2, f_3, f_4, f_5, f_{11}, f_{17}, f_{18}, f_{19}]$\newline
  {Note}: bookshelf correctly first-detected at $f_5$ (pre-glare, clean),
  not $f_{11}$ (where TV also appears). $f_{10}$ excluded as borderline
  flare ($(1{-}d)=0.39 < \theta_{\mathrm{rel}}$). \\
\myhdash

%% ── Tool Outputs ─────────────────────────────────────────────────────────────
%% Dominant corruption on selected frames = CLEAN (d_bar small).
%% Routing: detect_objects for all spatial sub-queries; track_temporal
%% for temporal ordering. rho(F) ~ 0.65 throughout (mild brightness
%% deviation); no HIGH tier but no contradictions among 5 MED-tier facts.
\lblcell{Tool Outputs}{\textit{(result, conf)}} &
  {\setlength{\tabcolsep}{2pt}%
  \renewcommand{\arraystretch}{1.35}%
  \begin{tabular}[t]{@{} p{3.0cm} @{\quad}p{2.9cm} p{2.7cm} r@{\,} c @{}}
    \specialrule{0.4pt}{0pt}{2pt}
    \textbf{Sub-query} & \textbf{Tool Call} & \textbf{Result} &
    \multicolumn{2}{l}{\textbf{Conf.\enspace Tier}} \\
    \specialrule{0.4pt}{1pt}{2pt}
    $sq_1$: \textit{Detect the yellow armchair} &
    \texttt{detect\_objects}\newline($f_1$, $f_2$, $f_3$) &
    First seen: \newline$f_2$ (score\,0.94) &
    0.611 & \texttt{MED} \\[3pt]
    $sq_2$: \textit{Detect the bookshelf with red books} &
    \texttt{detect\_objects}\newline($f_4$, $f_5$, $f_{11}$) &
    First seen: \newline$f_5$ (score\,0.89) &
    0.587 & \texttt{MED} \\[3pt]
    $sq_3$: \textit{Detect the wall-mounted television} &
    \texttt{detect\_objects}\newline($f_{11}$, $f_{17}$, $f_{18}$) &
    First seen: \newline$f_{11}$ (score\,0.92) &
    0.607 & \texttt{MED} \\[3pt]
    $sq_4$: \textit{Detect the houseplant in ceramic pot} &
    \texttt{detect\_objects}\newline($f_{17}$, $f_{18}$, $f_{19}$) &
    First seen: \newline$f_{18}$ (score\,0.95) &
    0.618 & \texttt{MED} \\[3pt]
    $sq_5$: \textit{Temporal order of objects} &
    \texttt{track\_temporal}\newline($f_1$--$f_5$, $f_{11}$,
    $f_{17}$--$f_{19}$) &
    $f_2\!\to\!f_5\!\to\!f_{11}\!\to\!f_{18}$\newline
    (A$\to$B$\to$C$\to$D) &
    0.592 & \texttt{MED} \\
    \specialrule{0.4pt}{1pt}{0pt}
  \end{tabular}} \\
\myhdash

%% ── Generated Answer ─────────────────────────────────────────────────────────
\textbf{Generated Answer} &
  {\textcolor{anscol}%
    {\textbf{(A)}~armchair $\to$ \textbf{(B)}~bookshelf $\to$
    \textbf{(C)}~television $\to$ \textbf{(D)}~houseplant.}
    Armchair at $f_2$, bookshelf at $f_5$ (pre-glare clean frame),
    television at $f_{11}$, houseplant at $f_{18}$, confirmed by
    \texttt{track\_temporal}. Glare $f_6$--$f_9$ and pet $f_{13}$--$f_{16}$
    gated out without loss of earliest-appearance evidence.
    (5~\texttt{MED}-tier; no contradictions; overall confidence:
    \texttt{MED}-leaning-\texttt{HIGH}).} \\
\myhdash

%% ── Ground Truth ─────────────────────────────────────────────────────────────
\textbf{Ground Truth} &
  \textbf{(A)}~armchair $\to$ \textbf{(B)}~bookshelf $\to$
  \textbf{(C)}~television $\to$ \textbf{(D)}~houseplant. \\
\bottomrule
\end{tabular}
\end{table}

%=====================================================================
% \section{Details of Parameter Setting}
%======================================================================
% APPENDIX SECTION: Details of Parameter Setting (single-paragraph form)
% Drop-in replacement for the empty \section{Details of Parameter Setting}.
%
% Note: The reward composition is reported strictly via Eq.(10) with the
% single auxiliary-reward weight w = 1/3, the definition given in §3.3.
% Table 17's per-term weights are NOT cited here, to avoid implying any
% (non-trivial) equivalence between the two formulations.
%======================================================================

%=====================================================================

%=====================================================================

%=====================================================================

\clearpage
\newpage
\section{Tool Invocation Statistics}
\label{appendix:tool_stats}

To understand how \methodname{} allocates its computational budget, we collect tool invocation statistics across the full evaluation set. \cref{tab:tool_invocation} reports the average number of calls per tool per question, broken down by clean and corrupted settings.

\begin{table}[t]
\centering
\small
\caption{Average tool invocations per question on UrbanVideo-Bench (Qwen2.5-VL-7B + \methodname{}). ``Clean'' denotes the original benchmark; ``Corrupted'' averages over all five RoVA masks.}
\label{tab:tool_invocation}
\begin{tabular}{@{}lcc@{}}
\toprule
\textbf{Tool} & \textbf{Clean} & \textbf{Corrupted} \\
\midrule
\texttt{assess\_quality}   & 1.00 & 1.00 \\
\texttt{select\_frames}    & 1.00 & 1.00 \\
\texttt{detect\_objects}   & 1.42 & 0.87 \\
\texttt{caption\_frame}    & 0.95 & 1.68 \\
\texttt{track\_temporal}   & 0.61 & 0.53 \\
\texttt{recognize\_action} & 0.48 & 0.72 \\
\texttt{read\_text}        & 0.32 & 0.18 \\
\texttt{retrieve\_frames}  & 0.08 & 0.24 \\
\midrule
Total calls per question   & 5.86 & 6.22 \\
Avg.\ sub-queries per question & 3.1 & 3.4 \\
Avg.\ selected frames $K$      & 7.8 & 6.2 \\
\bottomrule
\end{tabular}
\end{table}

\begin{table*}[t!]
\centering
\small
\caption{Learned routing preferences under dominant corruption modes. Each cell shows the tool preferred by the trained host VLM for a given sub-query type and corruption mode. The ``Clean'' column shows the default when no corruption dominates. \textit{Abbreviations:} \texttt{det\_obj}=\texttt{detect\_objects}, \texttt{cap}=\texttt{caption\_frame}, \texttt{trk\_tmp}=\texttt{track\_temporal}, \texttt{rec\_act}=\texttt{recognize\_action}, \texttt{rd\_txt}=\texttt{read\_text}.}
\label{tab:routing_rules}
\setlength{\tabcolsep}{10pt}
\begin{tabular}{@{}lcccc@{}}
\toprule
\textbf{Sub-query type} & \textbf{Clean} & \textbf{Blur} & \textbf{Brightness} & \textbf{Occlusion} \\
\midrule
Object identity / location   & \texttt{det\_obj} & \texttt{cap}     & \texttt{det\_obj} & \texttt{cap} \\
Object attribute / appearance & \texttt{cap}      & \texttt{cap}     & \texttt{cap}      & \texttt{det\_obj} \\
Temporal / motion             & \texttt{trk\_tmp} & \texttt{rec\_act}& \texttt{trk\_tmp} & \texttt{rec\_act} \\
Action / event                & \texttt{rec\_act} & \texttt{cap}     & \texttt{rec\_act} & \texttt{rec\_act} \\
In-video text                 & \texttt{rd\_txt}  & \texttt{cap}     & \texttt{rd\_txt}  & \texttt{rd\_txt} \\
\bottomrule
\end{tabular}
\end{table*}
Several patterns emerge. First, \texttt{assess\_quality} and \texttt{select\_frames} are called exactly once per question in both settings, confirming their role as fixed first-stage operations. Second, under corruption, \texttt{detect\_objects} calls decrease ($1.42 \to 0.87$) while \texttt{caption\_frame} calls increase ($0.95 \to 1.68$), reflecting the learned routing preference: detection degrades under blur and occlusion, so the agent shifts toward captioning. Third, the average number of selected frames $K$ drops from 7.8 to 6.2 under corruption, indicating the agent becomes more selective when fewer frames pass the reliability threshold. Fourth, \texttt{retrieve\_frames} usage increases under corruption ($0.08 \to 0.24$), suggesting the agent occasionally retrieves additional frames when its initial selection yields insufficient high-confidence evidence.
%=====================================================================

\paragraph{Confidence-Guided Tool Routing Rules.} 
The two-stage routing described in \cref{sec:pipeline} maps the semantic type and dominant corruption to a tool choice. \cref{tab:routing_rules} summarizes the learned routing preferences after GRPO training. The routing is soft - the host VLM selects tools via in-context reasoning conditioned on the disturbance profile, not via a hard-coded lookup table.
\section{Prompt Templates}
\label{appendix:prompts}

We provide the key prompt templates used by the host VLM at each stage of the \methodname{} pipeline. Angle brackets \texttt{\{...\}} denote dynamically filled placeholders.

\subsection{Sub-Query Decomposition Prompt}
\label{appendix:prompt_decompose}
\begin{tcolorbox}[colback=blue!5, colframe=blue!65!black, title=Sub-Query Decomposition, fonttitle=\bfseries, drop shadow=black!50!white, sharp corners, boxrule=0.8mm, breakable]
\textbf{[Task]}\\
You are an expert video analyst. Your task is to decompose a complex question about a video into a minimal set of atomic sub-queries. Each sub-query must target exactly one perceptual primitive and be answerable by a single visual tool call. Do not generate redundant sub-queries.\\

\textbf{[Decomposition Guidelines]}
\begin{enumerate}
\item Identify the distinct perceptual demands implied by the question (e.g., object localization, action recognition, attribute comparison, text reading).
\item For each demand, formulate exactly one atomic sub-query targeting a single visual primitive.
\item Assign a semantic type to each sub-query: one of \texttt{[spatial, temporal, attribute, action, text]}.
\item Minimize the total number of sub-queries --- each must be strictly necessary.
\end{enumerate}

\textbf{[Input]}
\begin{itemize}
\item Video context: \texttt{\{video\_description\}}
\item Disturbance profile of selected frames:\\
\quad blur=\texttt{\{avg\_blur\}}, brightness=\texttt{\{avg\_bright\}}, occlusion=\texttt{\{avg\_occl\}}
\item Question: \texttt{\{original\_query\}}
\end{itemize}

\textbf{[Output Format]}\\
Output a JSON list of atomic sub-queries. Only output the JSON --- no explanations, no justifications, and no extra text of any kind.
\begin{verbatim}
[
  {"sub_query": "<sub-query text>", "type": "spatial"},
  {"sub_query": "<sub-query text>", "type": "temporal"},
  ...
]
\end{verbatim}
\end{tcolorbox}
\subsection{Tool Routing Prompt}
\label{appendix:prompt_routing}

\begin{tcolorbox}[colback=blue!5, colframe=blue!65!black, title=Confidence-guided Tool Routing, fonttitle=\bfseries, drop shadow=black!50!white, sharp corners, boxrule=0.8mm, breakable]
\textbf{[Task]}\\
You are a tool routing agent. Given a sub-query, its semantic type, and the disturbance profile of the selected frames, choose the best perception tool from the available library that maximizes result reliability under the current corruption conditions.\\

\textbf{[Routing Guidelines]}
\begin{itemize}
\item For \textbf{spatial} sub-queries under blur: prefer \texttt{caption\_frame} over \texttt{detect\_objects} (detection requires sharp visual boundaries).
\item For \textbf{temporal} sub-queries under occlusion: prefer \texttt{recognize\_action} over \texttt{track\_temporal} (tracking loses targets under occlusion).
\item For \textbf{text} sub-queries under blur: prefer \texttt{caption\_frame} over \texttt{read\_text} (OCR degrades rapidly under spatial blur).
\item When brightness distortion dominates: prioritize tools robust to extreme illumination.
\item When multiple tools are viable: prefer the one with lower cost.
\end{itemize}

\textbf{[Input]}
\begin{itemize}
\item Sub-query: \texttt{\{sub\_query\_text\}}
\item Semantic type: \texttt{\{type\}}
\item Disturbance profile: blur=\texttt{\{d\_blur\}}, brightness=\texttt{\{d\_bright\}}, occlusion=\texttt{\{d\_occl\}}
\item Dominant corruption: \texttt{\{dominant\_type\}}
\item Available tools: \texttt{\{tool\_list\_with\_costs\}}
\end{itemize}

\textbf{[Output Format]}\\
Select the best tool. Only output the JSON --- no explanations beyond the reason field.
\begin{verbatim}
{
  "tool": "<tool_name>",
  "reason": "<one-sentence justification>"
}
\end{verbatim}
\end{tcolorbox}

\subsection{Confidence-Weighted Evidence Synthesis Prompt}
\label{appendix:prompt_synthesis}

\begin{tcolorbox}[colback=blue!5, colframe=blue!65!black, title=Confidence-Weighted Evidence Synthesis, fonttitle=\bfseries, drop shadow=black!50!white, sharp corners, boxrule=0.8mm, breakable]
\textbf{[Task]}\\
You are synthesizing evidence collected from multiple visual tools to answer a question about a video. Each piece of evidence has a confidence score (0--1) and a source frame disturbance level. Your goal is to produce a reliable answer grounded in the most trustworthy evidence.\\

\textbf{[Synthesis Rules]}
\begin{enumerate}
\item Group evidence into three reliability tiers:
  \begin{itemize}
  \item \textbf{HIGH}: confidence $\geq 0.7$ and disturbance $< 0.3$
  \item \textbf{MEDIUM}: all other evidence
  \item \textbf{LOW}: confidence $< 0.3$ or disturbance $\geq 0.7$
  \end{itemize}
\item Build your answer primarily from HIGH-tier evidence.
\item Use MEDIUM-tier evidence only if it is consistent with HIGH-tier conclusions; discard it if contradictory.
\item Use LOW-tier evidence only when no HIGH-tier evidence exists, and explicitly note the uncertainty.
\item If all evidence is LOW-tier, state that the answer is uncertain.
\end{enumerate}

\textbf{[Input]}
\begin{itemize}
\item Question: \texttt{\{original\_query\}}
\item Sub-queries and collected evidence:\\
\quad\textit{For each sub-query:}
  \begin{itemize}
  \item Sub-query: \texttt{\{sq\_text\}}
  \item Tool: \texttt{\{tool\_name\}}, Result: \texttt{\{result\}}, Confidence: \texttt{\{c\_j\}}
  \item Source frames: \texttt{\{frame\_ids\}}, Disturbance: \texttt{\{d\_scores\}}
  \end{itemize}
\end{itemize}

\textbf{[Output Format]}\\
Provide your step-by-step reasoning inside \texttt{<think>} tags, then your final answer inside \texttt{<answer>} tags. Only output these two blocks.
\begin{verbatim}
<think>
Step-by-step reasoning considering evidence reliability
and tier-based synthesis...
</think>
<answer>X</answer>
\end{verbatim}
\end{tcolorbox}

\subsection{$m^{*}$ Estimation Prompt ($\pi_{\mathrm{est}}$)}
\label{appendix:prompt_nstar}
\begin{tcolorbox}[colback=blue!5, colframe=blue!65!black, title=$m^{*}_q$ Sub-Query Count Estimation, fonttitle=\bfseries, drop shadow=black!50!white, sharp corners, boxrule=0.8mm, breakable]
\textbf{[Task]}\\
You are estimating the minimum number of independent perceptual sub-queries needed to fully answer a video question. This estimate is used as a reference target for training --- do not overestimate.\\

\textbf{[Estimation Guidelines]}
\begin{enumerate}
\item Count how many distinct objects, actions, or spatial relations the question asks about.
\item Determine whether temporal reasoning across multiple moments is needed (adds sub-queries).
\item Assess whether the question can be answered from a single frame or requires multi-frame evidence.
\item Each sub-query should be strictly necessary --- do not pad the count.
\end{enumerate}

\textbf{[Input]}
\begin{itemize}
\item Question: \texttt{\{original\_query\}}
\item Answer choices: \texttt{\{choices\_if\_multiple\_choice\}}
\end{itemize}

\textbf{[Output Format]}\\
Output a single integer representing the estimated number of atomic sub-queries needed. No explanations, no extra text.
\begin{verbatim}
<integer>
\end{verbatim}
\end{tcolorbox}

%=====================================================================

% \newpage
% \input{checklist.tex}

\end{document}